\RequirePackage{fix-cm}
\documentclass[twocolumn]{svjour3}          % twocolumn
\smartqed  % flush right qed marks, e.g. at end of proof
\usepackage{booktabs}
\usepackage{epsfig}
\usepackage{subfigure}
\usepackage{multirow}
\usepackage{array}
\usepackage{mathrsfs}
\usepackage{amsmath}
\usepackage{amssymb}
\usepackage{bm}
\usepackage{enumerate}
\usepackage{graphicx}
\usepackage[utf8]{inputenc}
\usepackage[english]{babel}
\usepackage[round, sort]{natbib}

\usepackage{setspace}
\usepackage{algorithm}
\usepackage{algpseudocode}
\usepackage{lipsum}
\usepackage{hyperref}
\usepackage{color}
\usepackage{marvosym}

\begin{document}

\title{Learning Accurate Performance Predictors for Ultrafast Automated Model Compression%\thanks{Grants or other notes
}


\author{Ziwei~Wang \and
	Jiwen~Lu \and
	Han~Xiao \and 
	Shengyu~Liu \and
	Jie~Zhou
}


\institute{Ziwei Wang\textsuperscript{1} \at
              \email{wang-zw18@mails.tsinghua.edu.cn}   \\       %  \\
           Jiwen Lu\textsuperscript{1,\textrm{\Letter}} \at
              \email{lujiwen@tsinghua.edu.cn} \\
           Han Xiao\textsuperscript{1} \at
              \email{h-xiao20@mails.tsinghua.edu.cn} \\
           Shengyu Liu\textsuperscript{1} \at
              \email{liusheng17 @mails.tsinghua.edu.cn} \\
           Jie Zhou\textsuperscript{1} \at
              \email{jzhou@tsinghua.edu.cn}      \\ \\
           \textsuperscript{1} State Key Lab of Intelligent Technologies and Systems, Beijing National Research Center for Information Science and Technology (BNRist), Department of Automation, Tsinghua University, Beijing, 100084, China
           %\textsuperscript{2} State Key Lab of Intelligent Technologies and Systems, China\\
           %\textsuperscript{3} Beijing National Research Center for Information Science and Technology, China  \\ \\
           %Code: https://github.com/ZiweiWangTHU/SeerNet.git.     
}

\date{Received: date / Accepted: date}

\maketitle

\begin{abstract}
In this paper, we propose an ultrafast automated model compression framework called SeerNet for flexible network deployment. Conventional non-differen-tiable methods discretely search the desirable compression policy based on the accuracy from exhaustively trained lightweight models, and existing differentiable methods optimize an extremely large supernet to obtain the required compressed model for deployment. They both cause heavy computational cost due to the complex compression policy search and evaluation process. On the contrary, we obtain the optimal efficient networks by directly optimizing the compression policy with an accurate performance predictor, where the ultrafast automated model compression for various computational cost constraint is achieved without complex compression policy search and evaluation. Specifically, we first train the performance predictor based on the accuracy from uncertain compression policies actively selected by efficient evolutionary search, so that informative supervision is provided to learn the accurate performance predictor with acceptable cost. Then we leverage the gradient that maximizes the predicted performance under the barrier complexity constraint for ultrafast acquisition of the desirable compression policy, where adaptive update stepsizes with momentum are employed to enhance optimality of the acquired pruning and quantization strategy.  Compared with the state-of-the-art automated model compression methods, experimental results on image classification and object detection show that our method achieves competitive accuracy-complexity trade-offs with significant reduction of the search cost. Code is available at \href{https://github.com/ZiweiWangTHU/SeerNet}{https://github.com/ZiweiWangTHU/SeerNet}.

\end{abstract}

\begin{figure}[t]
	\centering
	\includegraphics[height=12.3cm, width=8.5cm]{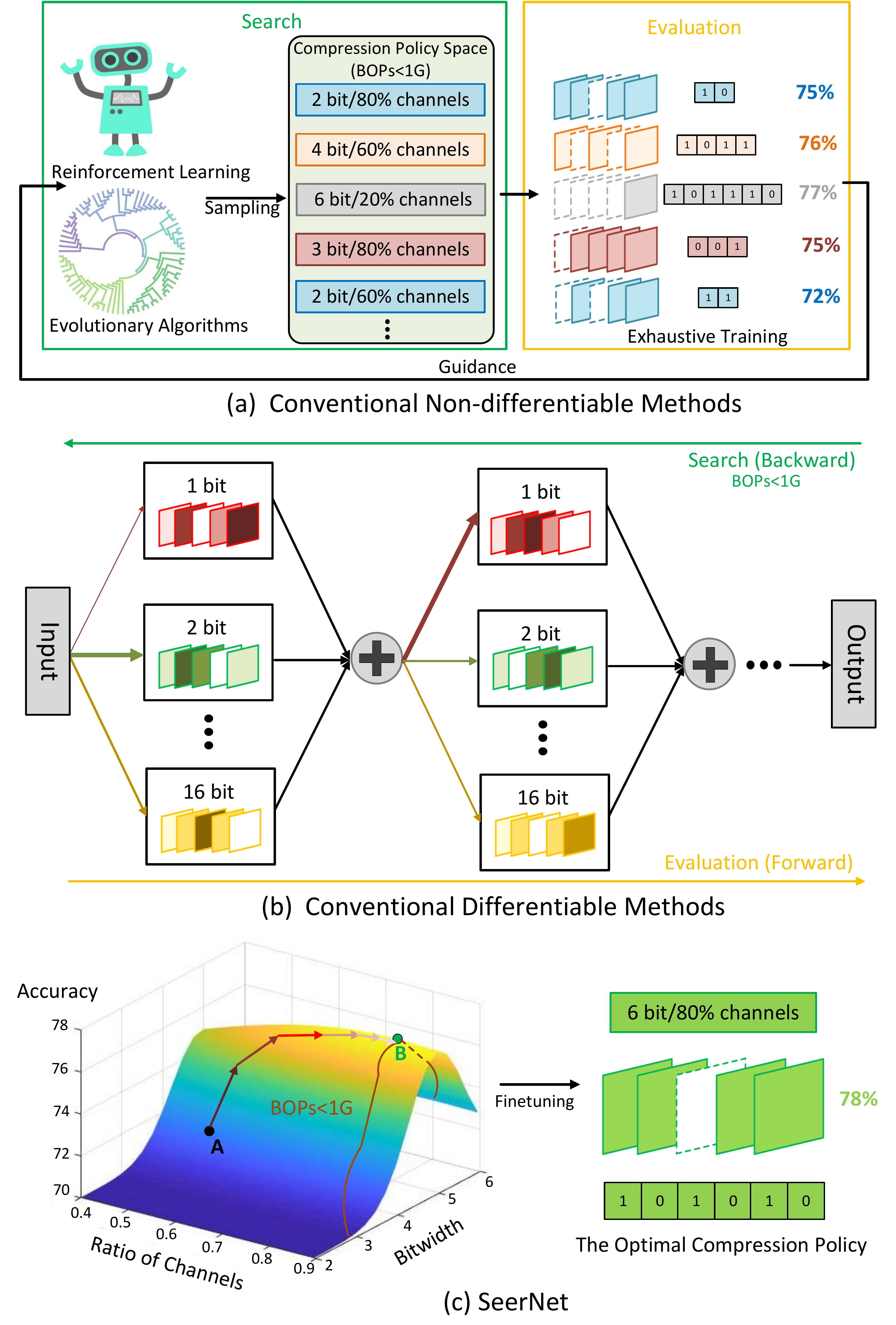}
	\caption{Comparison between (a) non-differentiable methods, (b) differentiable methods and (c) our SeerNet.Conventional non-differentiable and differentiable frameworks both result in heavy computational cost due to the complex compression policy search and evaluation stages. Our SeerNet directly optimizes the compression policy with a learned accurate performance predictor via efficient gradient ascent, and obtains the networks for deployment by finetuning the lightweight models compressed via selected pruning and quantization policies.}   
	\label{comparison}
	\vspace{-0.3cm}
\end{figure}

\section{Introduction}
\label{intro}
Deep neural networks have achieved the state-of-the-art performance on a wide range of vision tasks such as image classification \citep{he2016deep, simonyan2014very, phan2019mobinet}, object detection \citep{ren2015faster, liu2016ssd}, video analysis \citep{feichtenhofer2019slowfast, wang2019learning1} and many others. Nevertheless, deploying deep neural networks on mobile devices with limited resources for inference is usually impractical due to the heavy computational and storage complexity. Moreover, parameters in well-trained networks are proven to be highly redundant \citep{denil2013predicting}. Therefore, it is necessary to compress deep neural networks according to hardware configurations for flexible deployment.

In order to reduce the complexity of deep models, network pruning \citep{he2017channel, molchanov2019importance, liu2018rethinking} and quantization methods \citep{wang2021learning,wang2022learning,wang2022quantformer} have been widely studied, which also degrade the model performance due to the network capacity decreases. Pruning removes redundant model components that have little impact on performance, and quantization decreases the bitwidth of weights and activations with low-precision Multiply-Accumulate operations (MACs). Because the hardware resources vary across different deployment scenarios, selecting the optimal compression policy under the device constraint is important to obtain the ideal performance. Hardware equipment with strict resource limit should adopt extremely compressed models to satisfy the complexity constraint, while that with adequate resources only requires slight network complexity reduction to achieve high performance. To accomplish this, automated model compression methods have been proposed, where the optimal pruning ratio or the bitwidth for each convolutional layer is chosen according to the accuracy-complexity trade-off. Non-differentiable methods \citep{he2018amc, wang2019haq, lou2019autoq} applying reinforcement learning and evolutionary algorithms discretely search the optimal compression policy based on the accuracy from exhaustively trained lightweight models, and differentiable approaches \citep{wang2021generalizable, wang2020differentiable, qu2020adaptive} optimize the component weights in an extremely large supernet containing all compression policies to acquire the desired lightweight model for deployment. However, the complex compression policy search and evaluation process in both non-differentiable and differentiable methods leads to heavy search cost for automated model compression. For example, the hardware configurations of mobile devices can be selected from different GPUs, FPGAs and many others, and the battery levels can also vary during usage. Therefore, conventional methods with heavy search cost prohibit flexible network deployment due to the frequent changes of model complexity constraint. 

In this paper, we present an ultrafast SeerNet framework to learn the optimal model compression policy with the device resource constraint for flexible network deployment. Unlike existing non-differentiable and differentiable methods which undergo the complex compression policy search and evaluation process, our method directly optimizes the compression policy with an accurate performance predictor. The optimal compression policy is obtained via gradient ascent that maximizes the predicted accuracy, so that the efficiency of flexible model deployment is dramatically enhanced via removing resource-exhaustive compression policy search and evaluation. More specifically, we first learn the performance predictor via the accuracy from uncertain compression policies actively selected by evolutionary search, where the uncertainty is estimated via the performance variation with respect to the compression policy perturbation. The actively selected uncertain compression policies offer informative supervision to learn accurate performance predictor with acceptable cost, which can be utilized for flexible model deployment under different hardware scenarios. Then the gradient that maximizes the predicted accuracy under the barrier complexity constraint is leveraged for ultrafast acquisition of the desirable compression policy, and adaptive update stepsizes with momentum are utilized to strengthen the optimality of the obtained pruning and quantization strategy. Figure \ref{comparison} demonstrates the comparison between our SeerNet and the conventional non-differentiable and differentiable automated model compression methods with complexity constraint calculated by Bit-Operations (BOPs), where our framework achieves ultrafast compression policy selection for flexible network deployment. Compared with the state-of-the-art automated model compression methods, experiments on the CIFAR-10 \citep{krizhevsky2009learning} and ImageNet \citep{deng2009imagenet} for image classification and on PASCAL VOC \citep{everingham2010pascal} and COCO \citep{lin2014microsoft} for object detection show that our method achieves competitive performance with significantly reduced search cost. Our contributions are summarized as follows:
\begin{enumerate}[(1)]	
	\item We propose the ultrafast compression policy optimization framework which differentiably searches the pruning and quantization strategies on the performance predictor with the highest accuracy under the constraint of computational cost budget.
	
	\item We present an active compression policy evaluation method that samples the most uncertain pruning and quantization strategies, so that the accurate performance predictor is learned in acceptable training cost with informative supervision.
	
	\item We conduct extensive experiments on image classification and object detection, and the results consistently show that the presented SeerNet achieves competitive accuracy-complexity trade-offs with significant reduction of compression policy search cost.
	
\end{enumerate}

\section{Related Work}
We briefly review three related topics including (1) model compression, (2) AutoML and (3) active learning.

\subsection{Model Compression}
Pruning and Quantization are two widely adopted strategies for model compression. Pruning aims to remove the unimportant network components that have least influence on the performance, while quantization decreases the bitwidths of network weights and activations and substitutes float MACs with the low-precision ones.

Network pruning has been comprehensively studied in recent years because the model performance is nearly unaffected with sizable complexity degradation. Early attempts \citep{han2015deep, liu2015sparse} cut off the redundant fine-grained neurons and connections in an unstructured manner, which limited the actual acceleration on hardware equipment due to the irregular weight parameters. To address this, channel-pruning methods were later proposed, where the entire convolution channels were pruned according to the defined importance score. \cite{he2017channel} iteratively selected the channels for pruning with Lasso regression and finetuned the lightweight networks. The definition of the channel importance score have been also widely studied for effective pruning. The L1 and L2 norm of activations were used as the importance score in \citep{li2016pruning} and \citep{he2018soft} respectively. \cite{molchanov2016pruning} and \cite{peng2019collaborative} leveraged the first-order and second-order Taylor expansion with respect to the objective to evaluate the channel importance. Meanwhile, advanced sparsity regularization strategies \citep{louizos2017learning, li2019oicsr, li2020group} have been presented to achieve better trade-offs between the model accuracy and complexity. Nevertheless, the uniform pruning ratio across layers for various hardware configurations prohibits flexible network deployment due to the mismatch between hardware resources and model complexity.

Network quantization has been widely adopted in computer vision due to its efficiency in computation and storage, which is divided into one-bit and multi-bit quantization according to the bitwidth of network weights and activations. For the former, \cite{hubara2016binarized} and \cite{rastegari2016xnor} binarized weights and activations for efficient inference. \cite{liu2018bi} added an extra shortcut in consecutive layers to enhance the representational capacity of binary neural networks. \cite{gong2019differentiable} optimized the soft quantization strategy so that the discrepancy between the learning objective and the surrogate loss could be minimized. \cite{bethge2020meliusnet} increased the quality and capacity of features by channel enlargement and feature refinement, and created the efficient stem architectures to further reduce the computational cost of full-precision layers. Therefore, they even achieved higher accuracies than MobileNetV1 \citep{howard2017mobilenets} with similar computational complexity. Binary neural networks suffer from the extremely low network capacity, and multi-bit networks have been proposed with wider bitwidth and more sufficient representational power. \cite{choi2018pact} adaptively selected the activation clipping threshold to learn networks in 2-5 bits with high performance. \cite{zhang2018lq} minimized the quantization errors for all weights and activations to alleviate the information loss. \cite{li2019fully} overcome the training instabilities of four-bit object detectors with hardware-friendly implementations. Similar to pruning with uniform compression ratio, fixed-bit quantization cannot satisfy the demand of different deployment scenarios, where platforms with strict resource constraint require highly compressed models and vice versa.  

\subsection{AutoML}
Since the hardware configurations and battery levels vary significantly in different deployment scenarios, exploiting AutoML for automatic model compression arou-sed extensive interest in computer vision. The goal of AutoML is to select the compression policy that results in the best performance with the hardware resource constraint. Conventional AutoML frameworks for automatic model compression can be categorized into non-differentiable and differentiable methods based on the search strategy. For the former, \cite{he2018amc} and \cite{wang2019haq} applied the reinforcement learning to search the optimal layer-wise pruning and quantization policy respectively according to the accuracy from exhaustively trained networks. \cite{wang2020apq} used the evolutionary algorithms to acquire the desired compression policy and the architectures for the subnets of the once-for-all networks \citep{cai2019once}. For the latter, differentiable methods were presented to deal with the difficulties in discrete optimization of non-differentiable methods. \cite{cai2020rethinking} designed an extremely large supernet containing all quantization policies, and adjusted the importance of each quantization policy via back-propagation. \cite{wang2020differentiable} jointly searched the pruning and quantization policy via variational information bottleneck and the learned quantization mapping. \cite{yu2020search} constructed barrier penalty to ensure the obtained quantization policy satisfying the complexity constraint. However, complex model search and evaluation process in both non-differentiable and differentiable methods causes heavy computational cost for optimal compression policy acquisition, which prohibits the flexible network deployment for various hardware configurations and battery levels.  \cite{jin2020adabits} and \cite{bulat2021bit} trained a once-for-all network that could be quantized to any bits at runtime without finetuning. The once-for-all network quantization methods are orthogonal to automated model compression and can be combined with AutoML for further performance improvement.

\subsection{Active Learning}
Active learning enforces the model to acquire promising performance with few annotated training samples, where part of the training data providing effective supervision is labeled. The widely adopted criteria for annotation in active learning is based on the informativeness of the selected sample, which is evaluated by the prediction uncertainty. The uncertainty can be defined as the entropy of the posterior distribution \citep{joshi2009multi, luo2013latent, settles2008analysis}, disagreement among different classifiers \citep{melville2004diverse, vasisht2014active,wu2022smart}, difference between the largest and the second largest posterior probabilities \citep{balcan2007margin} and the distance to the boundary \citep{li2014multi, vijayanarasimhan2014large, abbasnejad2020counterfactual}. \cite{gal2017deep} employed deep neural networks to estimate task uncertainty through multiple forward passes in a data-driven manner. \cite{beluch2018power} presented a classifier committee to acquire accurate uncertainty estimation according to the disagreement. \cite{wang2020deep} selected informative samples for hash code learning by considering the pairwise similarity uncertainty. \cite{abbasnejad2020counterfactual} generated the most uncertain counterfactual sample with true labels by analyzing the performance sensitivity to the input perturbation. \cite{siddiqui2020viewal} measured the uncertainty of the semantic segmentation model via the inconsistency in predictions across viewpoints, which significantly lowered the cost of pixel-wise annotation. In this paper, we extend the active learning to efficiently train the accurate performance predictor with acceptable training cost, where only the most uncertain compression policy providing informative supervision is evaluated for actual accuracy acquisition.

\section{Approach}
In this section, we briefly review automatic model compression, which suffers from the heavy computational cost in compression policy search and evaluation. Then we introduce the details of compression policy optimization via the performance predictor. Finally, we propose active compression policy evaluation to learn the accurate performance predictor.

\subsection{Automated Model Compression}
The automated model compression is critical for deploying deep neural networks on different portable devices, as it provides the optimal compression policy with different computational cost constraint. The objective of automated model compression is written as follows:
\begin{align}
	\qquad\qquad\qquad~~ &\max\limits_{\bm{\mathcal{S}},\bm{\theta}} ACC_{val}(\bm{\mathcal{S}}(\bm{\mathcal{N}}), \bm{\theta})\notag\\
	&s.t. \quad C(\bm{\mathcal{S}}(\bm{\mathcal{N}}))\leqslant C_0
\end{align}where $\bm{\mathcal{N}}$ and $\bm{\mathcal{S}}$ are the original networks and the compression policy respectively. $\bm{\theta}$ represents the parameters of the compressed networks, and $ACC_{val}$ means the accuracy on the validation dataset. $C(\bm{\mathcal{S}}(\bm{\mathcal{N}}))$ stands for the complexity of the compressed networks and $C_0$ is the complexity constraint from device resources.

As shown in Figure \ref{comparison}(a), the non-differentiable methods take turns to search better compression policies and evaluate the sampled lightweight models. In the evaluation process, all sampled lightweight models are trained exhaustively to obtain the actual performance. During the search stage, agents in reinforcement learning or population in evolutionary algorithms are optimized to achieve higher accuracy with lower complexity, where the updated agents or population sample the best candidates for evaluation. As demonstrated in Figure \ref{comparison}(b), the differentiable methods optimize an extremely large supernet, where different compression policies form parallel modules for each layer. The output of all modules in each layer is added with different importance weights before being fed forward to the next layer. For the evaluation stage, the images are fed forward into the supernet to acquire loss value. For the search stage, importance weights of different modules are updated via back-propagation. The optimal compression policy is obtained by discretizing the soft module weights for the converged supernet.

However, compression policy search and evaluation in both non-differentiable and differentiable methods cause heavy computational cost. The mobile devices can be equipped with various hardware such as different GPUs, FPGAs and many others, and the battery levels can also vary during the usage. Hence, the heavy search cost prohibits flexible network deployment because of the frequent changes of model complexity constraint. Our goal is to remove the resource-exhaustive compression policy search and evaluation to achieve ultrafast automated model compression.
\vspace{-0.3cm}

\begin{figure}[t]
	\centering
	\includegraphics[height=6cm, width=8cm]{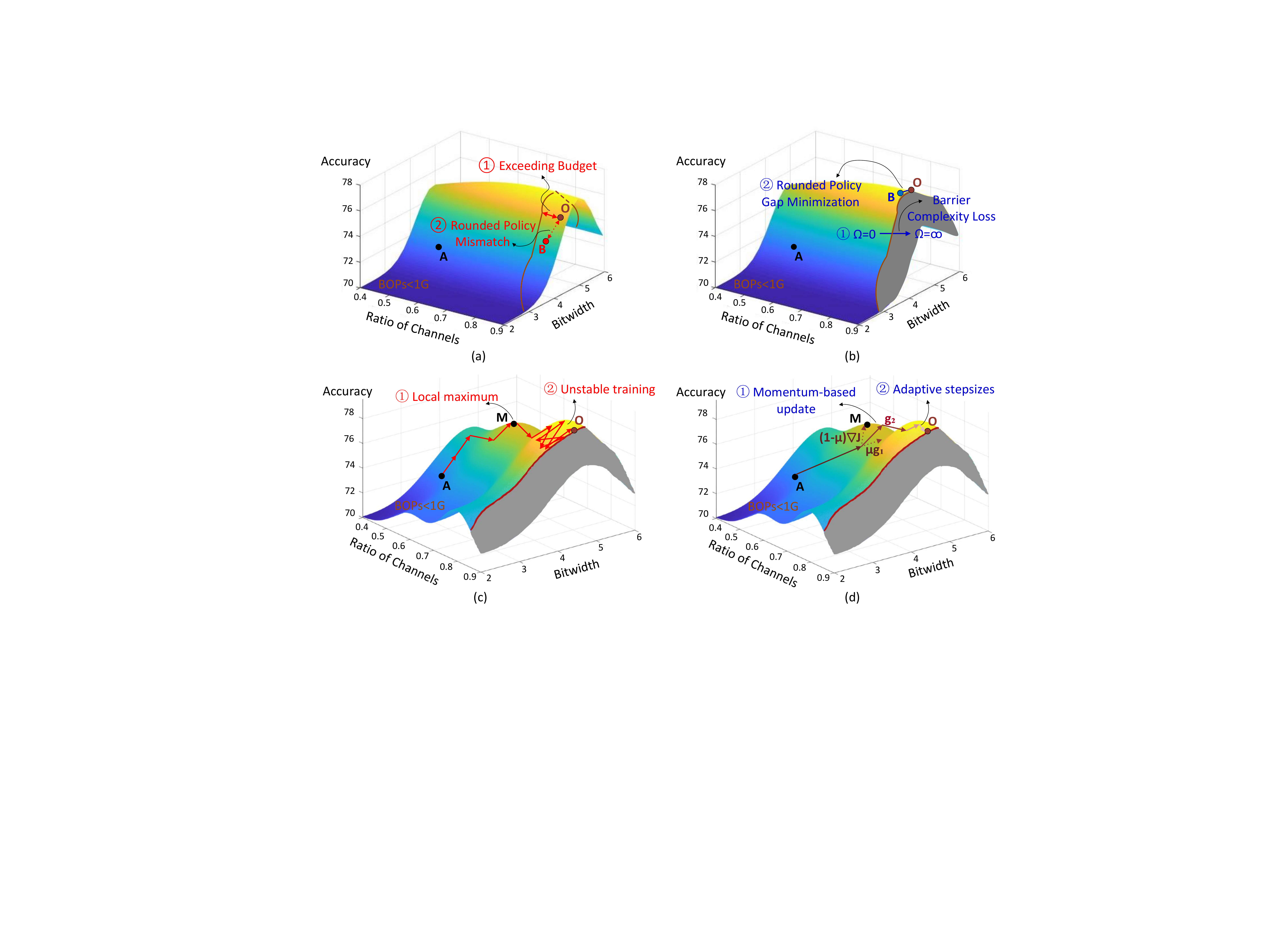}
	\caption{(a) and (b) visualize the learning objectives of (\ref{surrogate_obj}) and (\ref{reformulate_obj}) respectively, where A, O, B represent the initialized policy, the optimal policy after optimization and the discrete policy for deployment. Directly learning (\ref{surrogate_obj}) fails to rigidly limit the computational complexity of the lightweight model compressed by the searched policy within the budget due to the soft constraint, and results in sizable mismatch between the optimal policy and the rounded one for deployment because of the discrete nature of quantization and pruning policies. In order to address these problems, we present barrier complexity constraint and rounded policy gap minimization in the policy optimization objective. (c) and (d) demonstrate the optimization process of vanilla gradient ascent and (\ref{update}), where M means the local maximum. Vanilla gradient ascent faces the challenges of local maximum due to the non-convexity and unstable training process because of the fixed stepsizes. On the contrary, we present the momentum-based policy update to escape from the local maximum and propose adaptive stepsizes to stabilize the optimal policy search.}   
	\label{optimization}
	\vspace{-0.2cm}
\end{figure}

\subsection{Ultrafast Compression Policy Optimization}
In order to enhance the efficiency of automated model compression, we directly optimize the compression policy according to the learned performance predictor. In this section, we first introduce the learning objectives of compression policy optimization and then detail the compression policy update during the optimization.

\subsubsection{Learning Objectives}
The performance predictor consists of multi-layer perceptron (MLP), which takes the compression policies across all layers in the backbone architectures as input and predicts the accuracy of the lightweight models. Since the goal of automated model compression is to select the compression policy that leads to the highest accuracy with the given computational cost constraint, the objective $J$ for compression policy optimization is written in the following form:
\begin{align}
	\label{ultrafast_obj}
	\qquad\qquad\qquad\qquad\quad &\max J=f(\bm{s})\notag\\
	s.t. &\quad C(\bm{\mathcal{N}_s})\leqslant C_0
\end{align}where $f(\bm{s})$ means the predicted accuracy of the light-weight model with the compression policy $\bm{s}$. The definition of compression policy is $\bm{s}=[s_p^1, s_w^1, s_a^1,..., s_p^L, s_w^L, s_a^L]$, where $s_p^i$, $s_w^i$ and $s_a^i$ stand for the channel pruning ratio, weight bitwidth and activation bitwidth of the $i_{th}$ layer out of $L$ layers. In our implementation, weight bitwidth $s_w^i$ and activation bitwidth $s_a^i$ are scaled to $\frac{s_w^i}{s_{w,max}}$ and $\frac{s_a^i}{s_{a,max}}$, where $s_{w,max}$ and $s_{a,max}$ respectively represent the largest weight and activation bitwidth in the search space of compression policy. $\bm{\mathcal{N}_s}$ means the lightweight models obtained by compressing the original networks $\bm{\mathcal{N}}$ with the compression strategy $\bm{s}$. The network complexity is defined as Bit-Operations (BOPs) \citep{wang2020differentiable, bethge2020meliusnet, louizos2018relaxed} calculated in the following:
\begin{align}
	\label{complexity}
	\qquad ~~C(\bm{\mathcal{N}_s})=\sum_{i=1}^{L} s_w^i s_a^i (1-s_p^{i-1})(1-s_p^i)\cdot C_{ori}^i
\end{align}where $C_{ori}^i=h_{i}w_{i}k_h^ik_w^ic_{i-1}c_ib_w^ib_a^i$ demonstrates the BOPs of the $i_{th}$ layer in the original networks. $h_i$, $w_i$ and $c_i$ respectively represent the height, width and the number of channels of the output feature map in the $i_{th}$ layer, and $k_h^i$ and $k_w^i$ stand for the kernel height and width in the $i_{th}$ convolutional layer. For the full-precision networks, the weight and activation bitwidths of the $i_{th}$ layer denoted as $b_w^i$ and $b_a^i$ are usually set as $32$. Since BOPs reveal the effect of model complexity decrease induced by network pruning and quantization, we utilize the reduction ratio of BOPs to reflect the compression ratio.

Because better performance is usually obtained by networks with higher capacity, the optimal compression policy for (\ref{ultrafast_obj}) can be obtained when the model complexity achieves the computational cost constraint $C_0$. In order to efficiently optimize the desirable compression policy, the Lagrange multipliers can be employed to form the surrogate objective function with the hyperparameter $\lambda$, which is shown in the following:
\begin{align}
	\label{surrogate_obj}
	\qquad\qquad\qquad~ \max J=f(\bm{s})-\lambda C(\bm{\mathcal{N}_s})
\end{align}When the optimization completes, rounding the policy to the nearest one on grids yields the pruning and quantization policy for deployment due to their discrete nature. Nevertheless, directly optimizing (\ref{surrogate_obj}) deviates the obtained compression policy from the optimal one due to the following two reasons. First, the soft model complexity constraint in the objective cannot strictly limit the computational cost of the lightweight networks within the budget, which usually leads to suboptimal policies and huge search cost due to the repeated trials. Second, because the final compression policy for deployment is acquired by rounding the optimal policy to the nearest one on grids, the mismatch between the searched optimal policy and the discrete policy for deployment decreases the accuracy of the lightweight models. In order to address these problems, we formulate the objective function $J^{*}$ for compression policy optimization containing the barrier complexity constraint and rounded policy gap minimization:
\begin{align}
	\label{reformulate_obj}
	\qquad \max J^{*}=f(\bm{s})-\lambda_1 \Omega(C(\bm{\mathcal{N}_s}))-\lambda_2d(\bm{s},\bm{s}_0)
\end{align}where $\lambda_1$ and $\lambda_2$ are hyperparameters that demonstrate the importance of different objective terms. $\Omega(C(\bm{\mathcal{N}_s}))$ means the barrier complexity loss for the lightweight networks $\bm{\mathcal{N}_s}$, which is assigned to zero for $C(\bm{\mathcal{N}_s})$ less than $C_0$ and set to infinity otherwise. $d(\bm{s},\bm{s}_0)$ represents the distance between the compression policy $\bm{s}$ and its discrete counterpart $\bm{s}_0$. Figure \ref{optimization} (a) and (b) visualize the learning objectives of (\ref{surrogate_obj}) and (\ref{reformulate_obj}) respectively. The barrier complexity loss in our SeerNet ensures the obtained optimal compression policy to satisfy the computational budget with full utilization of computational resources, and the rounded policy gap is minimized to decrease the performance drop for policy discretization.

In order to enable the barrier complexity loss to be differentiable, we design $\Omega(C(\bm{\mathcal{N}_s}))$ with the following log-like function adopted from \citep{yu2020search, finlay2019logbarrier}:
\begin{align}
	\qquad\qquad \Omega(C(\bm{\mathcal{N}_s}))=-\log (C_0-C(\bm{\mathcal{N}_s}))
\end{align}Since $\Omega(C(\bm{\mathcal{N}_s}))$ is only rapidly amplified by the logarithm when approaching the complexity constraint $C_0$, the obtained lightweight model is strictly limited by the complexity constraint with full utilization of the computational resources. For the distance between the compression policy $\bm{s}$ and its discrete counterpart $\bm{s}_0$, we present the $L_2$ norm to measure their similarity, which is written as:
\begin{align}
	\qquad\qquad\qquad~~~ d(\bm{s},\bm{s}_0)=||\bm{s}-\bm{s}_0||_2^2
\end{align}where $||\cdot||_2$ represents the $L_2$ norm. Because $\bm{s}_0$ is obtained by rounding $\bm{s}$ to its nearest policy on grids, we relax $\bm{s}_0$ as a constant \citep{erin2015deep} for gradient back-propagation.

\setlength{\textfloatsep}{5pt}
\begin{algorithm}[t]
	\small
	\begin{spacing}{1.1}
		\begin{algorithmic}
			\Require 
			Network architecture $\mathcal{N}$, performance predictor $f$, computational cost constraint $C_0$, compression policy update round $Max\_iter$.
			\Ensure
			The optimal compression policy $\bm{s}^{*}$.
			\State \textbf{Initialize:} Randomly assign $\bm{s}$ where $C(\mathcal{N}_{\bm{s}})\approx\frac{C_0}{2}$. 
			\For {$t=1, 2, ..., Max\_iter$}
			\State Calculate the objective function of $\bm{s}_t$ via (\ref{reformulate_obj}).
			\State Compute the gradient with respect to $\bm{s}_t$ by (\ref{gradient}).
			\State Update the compression policy with the above gradient according to (\ref{update}).
			\If{$C(\mathcal{N}_{\bm{s}_t})\geqslant C_0$} 
			\Return compression policy $\bm{s}_{t}$.
			\EndIf
			\EndFor
			\State\Return compression policy $\bm{s}_{t+1}$.
		\end{algorithmic}
	\end{spacing}
	\caption{Compression policy optimization}
	\label{alg_ucpo}
\end{algorithm}

\subsubsection{Compression Policy Update}
As the predicted accuracy and the complexity of compressed models can both be obtained by the differentiable calculation, we leverage the gradient that maximizes the objective (\ref{reformulate_obj}) with momentum to update the compression policy:
\begin{align}
	\label{update}
	\qquad\qquad\qquad~~ \bm{s}_{t+1}=\bm{s}_{t}+\epsilon_t\cdot\frac{\bm{g}_t}{||\bm{g}_t||_2}
	%\epsilon\frac{\partial J}{\partial \bm{s}_{t}}
\end{align}where $\bm{s}_{t}$ means the compression policy in the $t_{th}$ step during the optimization. $\bm{g}_t$ illustrates the accumulated gradient in the $t_{th}$ step, and $\epsilon_t$ is defined as the stepsize in the $t_{th}$ step which is adaptively assigned. As indicated in \citep{dong2018boosting} that integrating the momentum into iterative processes of input update can boost optimization, we adopt the accumulated gradients in the following that escape from the local maximum \citep{duch1998optimization, sutskever2013importance}:
\begin{align}\label{gradient}
	\qquad\qquad~~ \bm{g}_{t+1}=\mu\cdot\bm{g}_t+(1-\mu)\cdot\frac{\nabla_{\bm{s}} J^{*}}{||\nabla_{\bm{s}} J^{*}||_2}
\end{align}where $\mu$ is a hyperparameter that balances the momentum and the current gradient in the accumulated gradients. In order to stabilize the training process \citep{kingma2014adam}, the stepsize for compression policy update in each step should be adjusted with respect to the complexity difference between the current lightweight model and the computational complexity constraint. When the current policy is far from the complexity constraint, the stepsize should be large in order to accelerate training. On the contrary, the stepsize should be small for policy optimization near the computational cost budget due to the extremely large barrier complexity loss, so that fine-grained optimization is adopted to stably search the optimal policy within the complexity constraint. We present the adaptive stepsize at the $t_{th}$ step as follows:
\begin{align}\label{ada_stepsize}
	\qquad\qquad\qquad ~~~ \epsilon_t=\eta\cdot(C_0-C(\bm{\mathcal{N}}_{\bm{s}_t}))
\end{align}where $\eta$ is a hyperparameter and $C(\bm{\mathcal{N}}_{\bm{s}_t})$ demonstrates the complexity of the lightweight models compressed by the policy in the $t_{th}$ iterative update step. Figure \ref{optimization} (c) and (d) illustrate the vanilla gradient ascent and the presented compression policy optimization respectively, where our optimization process escapes from the local maximum and stably obtains the policy with the highest accuracy within the complexity constraint.

The compression policy update process stops until reaching the computational cost constraint or achieving the maximum iteration steps. The detailed procedures of ultrafast compression policy optimization are shown in Algorithm \ref{alg_ucpo}, where flexible deployment across different hardware configurations and battery levels is achieved since the gradient of the performance predictor consisting of several MLPs is calculated with extremely little computational cost. 

\subsection{Learning Performance Predictor via Active Compression Policy Evaluation}
The acquisition of the optimal lightweight model via differentiable compression policy optimization requires the learned performance predictor to be precise, where the gap between the predicted and actual performance is negligible. Conventional accuracy predictors for network architecture search \citep{dai2019chamnet, wen2020neural} randomly sample compression policies, and acquire the actual performance by exhaustively training the lightweight models. Then the actual accuracy is employed to supervise the performance predictor that regresses the accuracy of the compression policy. However, the number of sampled lightweight models for evaluation is extremely small compared with the large space of compression policies due to the limited computational resources. Randomly sampled compression policies fail to provide informative supervision for performance predictor learning. On the contrary, we actively select the uncertain compression policy for evaluation to obtain its actual accuracy, and train the performance predictor with the sampled policy that offers informative supervision. We first demonstrate the performance predictor learning with policy uncertainty, and then depict the active selection for uncertain policy.

\begin{figure}[t]
	\centering
	\includegraphics[height=5.3cm, width=8cm]{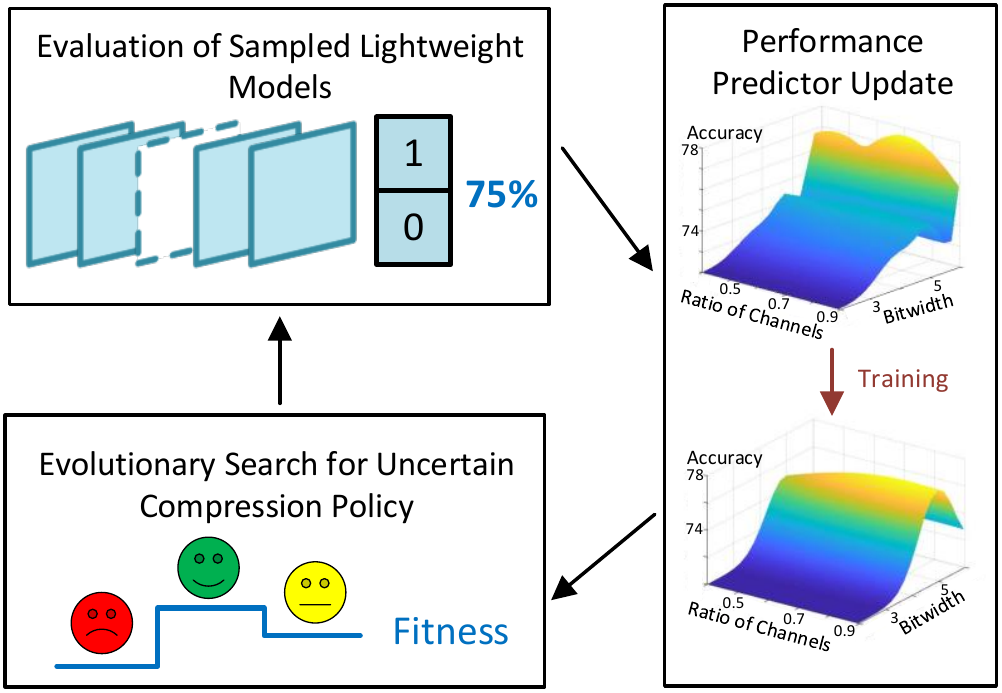}
	\caption{The pipeline of learning the performance predictor via active compression policy evaluation, where we iteratively search the most uncertain compression policy defined by (\ref{fitness}) via evolutionary algorithms, obtain the actual accuracy of sampled lightweight models via exhaustive training and update the performance predictor with the actual accuracy of sampled compression policy according to (\ref{alternative}).}   
	\label{pipeline}
\end{figure}

\subsubsection{Performance Predictor Learning with Policy Uncertainty}
Training the performance predictor via compression policies with uncertain prediction provides informative supervision, since the performance predictor obtains more knowledge from those samples \citep{beluch2018power, gal2017deep}. Therefore, exhaustively training models compressed by those policies makes significant contribution to enhance the precision of the performance predictor. Figure \ref{pipeline} illustrates the pipeline of performance predictor learning in our SeerNet. For a given backbone, we iteratively search the uncertain compression policy via evolutionary algorithms, evaluate the sampled lightweight models to obtain the actual accuracy, and update the performance predictor with the actual accuracy of sampled compression policies. The well-trained performance predictor is employed for ultrafast compression policy optimization, so that flexible network deployment with different resource constraint is achieved without complicated compression policy search and evaluation.

The influence of the compression policy perturbation on predicted performance reveals prediction uncertainty, where that sensitive to perturbation indicates highly uncertain prediction \citep{vijayanarasimhan2014large, abbasnejad2020counterfactual}. Hence, the training loss of more uncertain compression policies should be weighted more greatly to strengthen the supervision informativeness. We employ the importance sampling by reweighting samples in the objective function $R(\bm{w})$ to train the performance predictor with the parameters $\bm{w}$ \citep{abbasnejad2020counterfactual, goyal2019counterfactual}:
\begin{align}	\label{importance}
	~~\min\limits_{\bm{w}}R(\bm{w})&=\mathbb{E}_{\bm{s}\sim p(\bm{s})}\mathbb{E}_{a\sim p(a|\bm{s})}l(f(\bm{s}),a)\notag\\
	&=\mathbb{E}_{\bm{s}\sim p(\bm{s})}\mathbb{E}_{a\sim p(a|\bm{\hat{s}})}l(f(\bm{s}),a)\frac{p(a|\bm{s})}{p(a|\bm{\hat{s}})}	
\end{align}where $a$ means the actual accuracy and $\bm{\hat{s}}$ represents the perturbed counterpart of $\bm{s}$. $p(\bm{s})$ is the prior distribution of the compression policy. $p(a|\bm{s})$ and $p(a|\bm{\hat{s}})$ demonstrate the posterior distribution of accuracy given the compression policy $\bm{s}$ and the perturbed one $\bm{\hat{s}}$ respectively. $l(f(\bm{s}),a)$ is the loss function of accuracy prediction, which is defined as the mean squared error (MSE). In the importance sampling, the compression policy whose accuracy varies more significantly with the perturbation acquires larger weights in the learning objective. Since we leverage deterministic neural networks to predict the accuracy of various lightweight models, we optimize the following alternative objective $R^{*}(\bm{w})$ for the performance predictor, which is mathematically formulated in the appendix. The goal of (\ref{importance}) is to heavily weight the compression policy whose predicted accuracy is very different from the perturbed one, and we present the $L_2$ difference between predicted accuracies of the vanilla compression policy and the perturbed one as importance weights in the alternative objective:

\begin{small}
	\begin{align}\label{alternative}
		\min\limits_{\bm{w}} R^{*}(\bm{w})=\sum_{i=1}^{N}\sum_{\bm{\hat{s}}^i}(f(\bm{s}^i)-a^i)^2\cdot ||f(\bm{s}^i)-f(\bm{\hat{s}}^i)||_2^2
	\end{align}
\end{small}where $N$ is the number of actively sampled compression policies for performance predictor training. $\bm{s}^i$ and $\bm{\hat{s}}^i$ mean the $i^{th}$ sampled compression policy and its perturbed counterpart. $a^i$ represents the actual accuracy of the $i_{th}$ compressed model obtained via exhaustively training. The $L_2$ difference of the predicted accuracy between the compression policy $\bm{s}^i$ and the perturbed counterparts reflects the importance weight. By penalizing the compression policy that is more sensitive to perturbation, the accurate performance predictor is learned by informative supervision with acceptable training cost.

\begin{algorithm}[t]
	\small
	\begin{spacing}{1.1}
		\begin{algorithmic}
			\Require 
			Backbone Network $\mathcal{N}$, the number of compression policy sampling $K$, performance predictor learning round $Max\_ro$, evolution round $Max\_iter$.
			\Ensure
			Accurate performance predictor $f^{*}$.
			\State \textbf{Initialize:} Randomly assign the weights of $f$.
			\For {$t=1, 2, ..., Max\_ro$}
			\State Randomly sample $K/Max\_ro$ compression policy $\bm{s}$.
			\For {$i=1, 2, ..., Max\_iter$}
			\State Generate perturbed compression policy $\hat{\bm{s}}$ via (\ref{perturbation}).
			\State Predict the performance $f(\bm{s})$ and $f(\hat{\bm{s}})$.
			\State Select the top-k compression policy with the highest fitness according to (\ref{fitness}).
			\State Mutation and crossover for the next generation $\bm{s}$.
			\EndFor
			\State Train and validate $\mathcal{N}_{\bm{s}}$ for actual performance.
			\State Train $f$ with the actual performance of $\bm{s}$ via (\ref{alternative}).
			\EndFor
			\State\Return the performance predictor $f$.
		\end{algorithmic}
	\end{spacing}
	\caption{Active performance predictor learning}
	\label{alg_appl}
\end{algorithm}

\subsubsection{Active Selection for Uncertain Policy}
In this section, we introduce the details for the search strategy of uncertain compression policies, which provides informative supervision for performance predictor learning. The uncertainty of the compression policy is evaluated by the accuracy sensitivity with respect to the perturbation on the policy space. The network capacity revealed by the complexity varies differently with the channel pruning ratio or the bitwidths of weights and activations across layers, because the pruning and quantization policies for different layers contribute diversely to the overall BOPs. Since the network capacity has significant influence on the model accuracy, the compression policy variation for uncertainty evaluation should enforce all perturbed counterparts to change the network capacity identically. Therefore, the uncertainty of different policies should be fairly estimated without the impacts of the network capacity variation.

Specifically, we generate each perturbed compression policy $\bm{\hat{s}}$ for the original one $\bm{s}$ by modifying one element with the following criteria:
\begin{align}\label{perturbation}
	~~~~~\{\bm{\hat{s}}\big||s_k-\hat{s}_k|=\alpha_m\cdot\mathbb{I}[k=m],~ k=1,2,...,3L\}
\end{align}where $s_k$ and $\hat{s}_k$ mean the $k_{th}$ element of $\bm{s}$ and $\bm{\hat{s}}$, and the indicator function $\mathbb{I}[x]$ equals to one for true $x$ and to zero otherwise. By varying $m\in\{1,2,...,3L\}$, we acquire $3L$ perturbed policies that increase and decrease BOPs respectively, which result in $6L$ perturbed policies in total for the uncertainty evaluation of $\bm{s}$. Meanwhile, $\alpha_m$ is the scale coefficient to ensure the complexity variation consistency for various perturbed policies: 
\begin{align}
	\qquad\qquad\qquad ~~~\alpha_m=B_0\cdot (\frac{\partial C(\bm{\mathcal{N}_s})}{\partial s_m})^{-1}
\end{align}where $B_0$ is a hyperparameter that demonstrates the model capacity variation for policy perturbation, and $\frac{\partial C(\bm{\mathcal{N}_s})}{\partial s_m}$ depicts the sensitivity of the model complexity $C(\bm{\mathcal{N}_s})$ defined in (\ref{complexity}) with respect to the element $s_m$.

Because the compression policy with high prediction uncertainty contributes significantly in performance predictor learning according to (\ref{alternative}), we sample the compression policies for actual performance acquisition via the following criteria in order to provide most informative supervision:
\begin{align}
	\label{fitness}
	\qquad\qquad~~~\bm{s}=\arg\max \sum_{\bm{\hat{s}}} ||f(\bm{s})-f(\bm{\hat{s}})||_2
\end{align}Since the compression policy space is extremely large, we present the evolutionary search to select the most uncertain compression policy for the performance predictor training. In the evolutionary search, the genes are the vectors $\bm{s}$ representing the compression policy. We first randomly select the genes for initialization and obtain the their fitness $\mathcal{F}$ defined as $\mathcal{F}=\sum_{\bm{\hat{s}}} ||f(\bm{s})-f(\bm{\hat{s}})||_2$. The top-k genes with the highest fitness are chosen for generating off-spring genes via the mutation and crossover process. The mutation process is carried out by randomly varying a proportion of elements that demonstrate the pruning ratio and quantization bitwidths in the genes, and the crossover process means that we recombine the pruning and quantization policies in two parent genes for off-spring generation. By iteratively selecting the top-k genes with the highest fitness and generating new genes with mutation and crossover, the compression policy with the most uncertain prediction is selected, which offers informative supervision for performance predictor learning. As the fitness of candidates can be evaluated by predicting the accuracy of the compression policies and their perturbed counterparts, the evolutionary search for the uncertain compression policies is computationally efficient. Algorithm \ref{alg_appl} demonstrates the active performance predictor learning process, where the performance predictor is learned offline and utilized in ultrafast compression policy optimization for flexible deployment.

\iffalse
Because an accurate performance predictor enhances the precision of the fitness function in the evolutionary search, we sample the uncertain compression policy in several rounds. In each round, we iteratively optimize the performance predictor with the actual accuracy of models compressed by the sampled policy, and select the most uncertain compression policy via evolutionary algorithms for exhaustive evaluation that provides informative supervision.
\fi

\section{Experiments}
In this paper, we conducted extensive experiments to evaluate our methods on the CIFAR-10 and ImageNet datasets for image classification and on the PASCAL VOC and COCO datasets for object detection. We first briefly introduce the datasets and the implementation details, and then verify the effectiveness of the presented ultrafast compression policy optimization and active compression policy evaluation for performance predictor learning via ablation study. Finally, we compare our SeerNet with the existing automated model compression methods to show our superiority.

\subsection{Datasets and Implementation Details}
We introduce the datasets we carried experiments on and data preprocessing techniques in the following:

\textbf{CIFAR-10: }The CIFAR-10 dataset includes $60,000$ samples with the resolution of $32\times 32$, which are equally divided into $10$ classes. We leveraged $50,000$ and $10,000$ images as the training and test sets respectively. We padded $4$ pixels on each side of the images and randomly cropped them into the size of $32\times32$. Moreover, we scaled and biased all pixels into the range $[-1,1]$.

\textbf{ImageNet: }ImageNet (ILSVRC2012) consists of approximately $1.2$ million training and 50K validation images collected from $1,000$ categories. Following the data preprocessing techniques of bias extraction applied in CIFAR-10, we randomly cropped a $224\times 224$ region from the resized image whose shorter side was $256$ during the training process. For inference, we adopted a $224\times 224$ center crop from the validation images.

\textbf{PASCAL VOC: }PASCAL VOC includes images from $20$ different classes. Our model is trained on the VOC 2007 and VOC 2012 trainval sets consisting of about $16$k images, and we evaluated our SeerNet on VOC 2007 test set containing around $5$k images. Following \citep{everingham2010pascal}, we employed the mean average precision (mAP) as our evaluation criterion.

\textbf{COCO: }The images in the COCO dataset were collected from $80$ different categories, and our experiments were conducted on the 2014 COCO object detection track. We trained our model with the combination of $80$k images from the training set and $35$k images selected from validation set (trainval35k \citep{bell2016inside}), and tested our SeerNet on the remaining minival validation set \citep{bell2016inside} including $5$k images. Following the standard COCO evaluation metric \citep{lin2014microsoft},  we apply the mean average precision (AP) for IoU $\in  \left[0.5 : 0.05 : 0.95\right]$ as the evaluation metric. We also report average precision with the IOU threshold $50\%$ and $75\%$ represented as AP$_{50}$ and AP$_{75}$ respectively. Moreover, the average precision of small, medium and large objects notated as AP$_{s}$, AP$_{m}$ and AP$_{l}$ are also depicted.

For image classification, we employed architectures of VGG-small \citep{zhang2018lq} and ResNet20 \citep{he2016deep} for automated model compression on CIFAR-10, and compressed ResNet18, ResNet50 and MobileNetV2 \citep{sandler2018mobilenetv2} architectures with various computational cost constraint on ImageNet. For object detection, we adopted the SSD \citep{liu2016ssd} framework with VGG16 and the Faster R-CNN \citep{ren2015faster} framework with ResNet18. Our performance predictor consisted of three fully-connected layers with the ReLU activation function. We iteratively trained the performance predictor with the accuracy of sampled lightweight models and actively searched uncertain compression policies via evolutionary algorithms. 

In the sampled compression policies, the choices for the pruning ratio of all layers were $\{0.25, 0.5, 0.75\}$, while the selections for the weight and activation bitwidths of each layer were set as $\{2,4,6,8\}$. Since quantizing the weights and activations in the first and last layers with low-precision significantly degrade the model performance, we set the bitwidth of the first and last layers of the backbone to $8$ following \citep{wang2019haq}. For perturbed compression policy generation, the hyperparameter $B_0$ was positively related to the original BOPs of the full-precision networks. Varying each element in the compression policies yields $6L$ perturbation for networks with $L$ layers, and we only randomly sampled $0.5L$ perturbed compression policies for performance prediction to reduce the computational cost in uncertainty estimation. We trained $800$ lightweight models with different compression policies for accuracy acquisition in order to learn the performance predictor of each backbone and dataset. $50$ compressed models were actively sampled for evaluation and performance predictor training in each round out of $16$ rounds, where the compression policies that initially trained the performance predictor were randomly selected. We set the population size to be $100$ in the evolutionary search for uncertain compression policies, where the top-$25$ candidates based on (\ref{fitness}) produced the next generation. $50$ candidates randomly mutated with the mutation rate $0.1$ for the compression policy of each layer. For crossover, the compression policy of each layer was randomly chosen from $50$ parent candidates. The max iterations were $500$ for the best candidate selection. 

In compression policy optimization, the hyperparameters $\lambda_1$ and $\lambda_2$ in the objective were $0.1$ and $0.005$, and the hyperparameters $\mu$ for gradient accumulation and $\eta$ for adaptive stepsizes were $0.9$ and $0.05$. The maximum iteration step for updating the compression policy was $30$. We randomly selected compression policy whose complexity was approximately half of the computational cost constraint for initialization, and updated the compression policies until reaching the maximum iteration or the computational cost constraint.

We employed the max response selection \citep{han2015learning} that pruned weights according to the magnitude for channel pruning. Meanwhile, we followed the implementation in \citep{wang2019haq} for weight and activation quantization.
\iffalse
We linearly quantized the real-valued weight $w$ with the quantization bit $b$ into $[-v,v]$:
\begin{align}
	w^{'}=\max(0,\min(2v,round(\frac{2w}{2^b-1})\cdot v))-v
\end{align}where $w^{'}$ is the quantized weight and $v$ is chosen differently for various layers to minimize the quantization error. The activation was quantized in a similar way except the range of quantization was set as $[0,v]$ due to the ReLU layers. 
\fi
During training of the lightweight networks, we used the Adam optimizer \citep{kingma2014adam} with the batchsize of $256$. For CIFAR-10, we initialized the learning rate as $0.001$ and decayed twice at the $60_{th}$ and $80_{th}$ training epochs out of $100$ epochs, where the learning rate multiplied $0.1$ for each decay. For ImageNet, the learning rate started from $0.005$ and decayed at the $20_{th}$ and $40_{th}$ in the total $60$ epochs with the same decay rate. The backbone for object detection was pretrained on ImageNet following the above implementation details. For the network finetuning on object detection, the learning rate was initially set as $1e$-$3$ and decreased to $1e$-$4$ and $1e$-$5$ at the $40_{th}$ and $60_{th}$ epoch out of $80$ epochs for PASCAL VOC, and started from 0.001 with the same decay strategy at the $6_{th}$ and $10_{th}$ epoch during $12$ training epochs for COCO.

\subsection{Ablation Study}
To verify the benefits of active compression policy evaluation for performance predictor learning, we conducted the ablation study to assess our performance predictor w.r.t. different sampling strategies for compression policies and various numbers of sampled lightweight models on ResNet20 and ResNet50. With the same architectures, we varied the perturbation magnitude in uncertainty estimation with different numbers of perturbed compression policies in order to show the influence.

For the ablation study of compression policy optimization with ResNet20, we validate the effectiveness and efficiency by comparing the accuracy-complexity trade-off with the optimal lightweight models obtained via other search strategies including reinforcement learning and evolutionary algorithms. In order to verify the impact of the barrier complexity loss, the rounded policy gap minimization, the momentum-based policy update and adaptive stepsizes in our ultrafast compression policy optimization, we report the accuracy-complexity trade-off of compression policies obtained via different combinations of the above techniques. Moreover, we investigate the impact of the initialization and the stepsize scale of policy update. The ablation study was conducted on CIFAR-10 with the BOPs constraint $0.2$G and $0.4$G for ResNet20 and ResNet50 respectively.

\begin{figure}
	\centering
	\begin{center}
		\vspace{-0.4cm}
		\subfigure[ResNet20]{
			\begin{minipage}[b]{0.44\linewidth}\label{fig:Res20}
				\includegraphics[height=3.3cm, width=4.5cm]{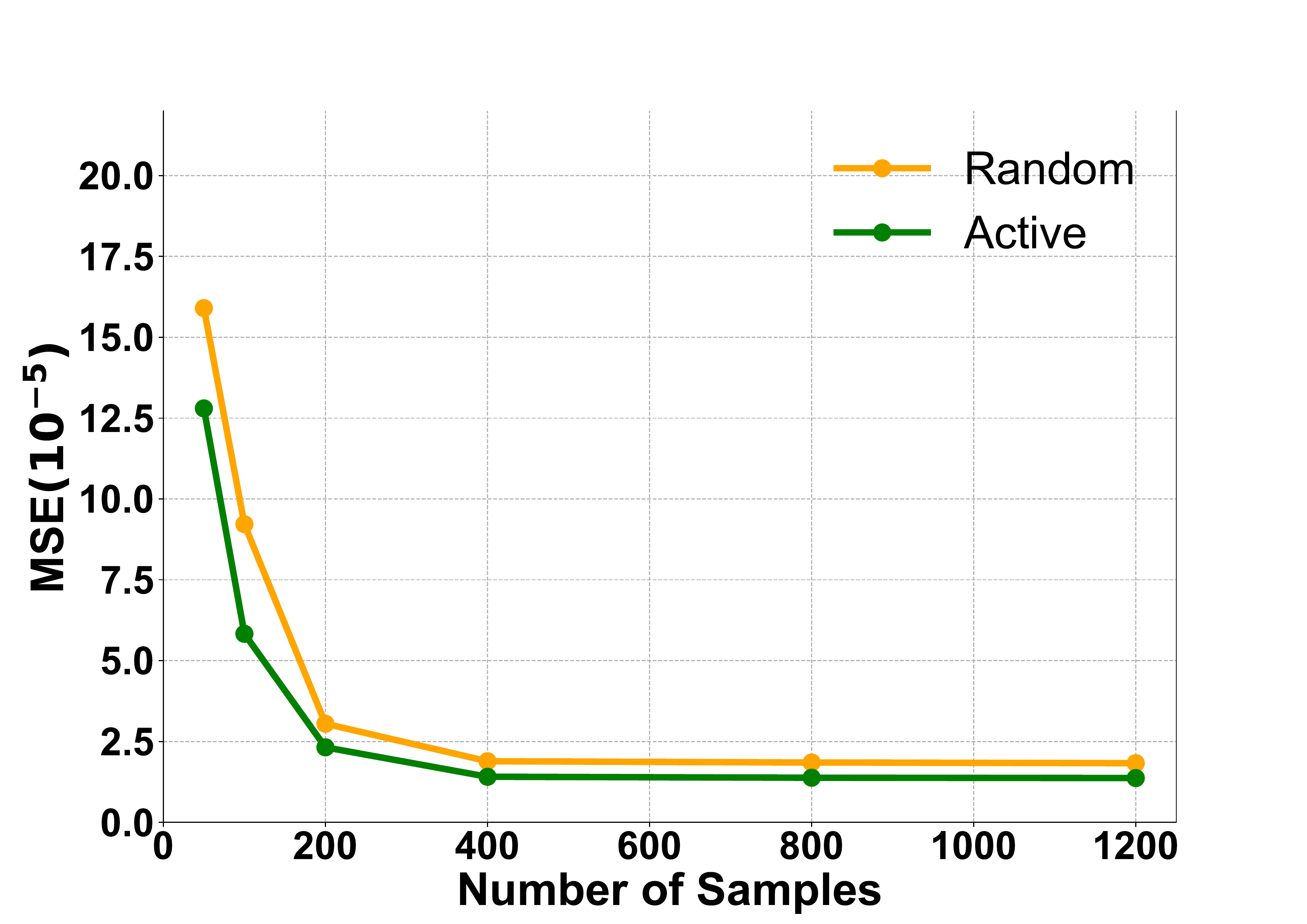}
			\end{minipage}
		}
		\hfill
		\subfigure[ResNet50]{
			\begin{minipage}[b]{0.47\linewidth}\label{fig:Res50}
				\includegraphics[height=3.3cm, width=4.5cm]{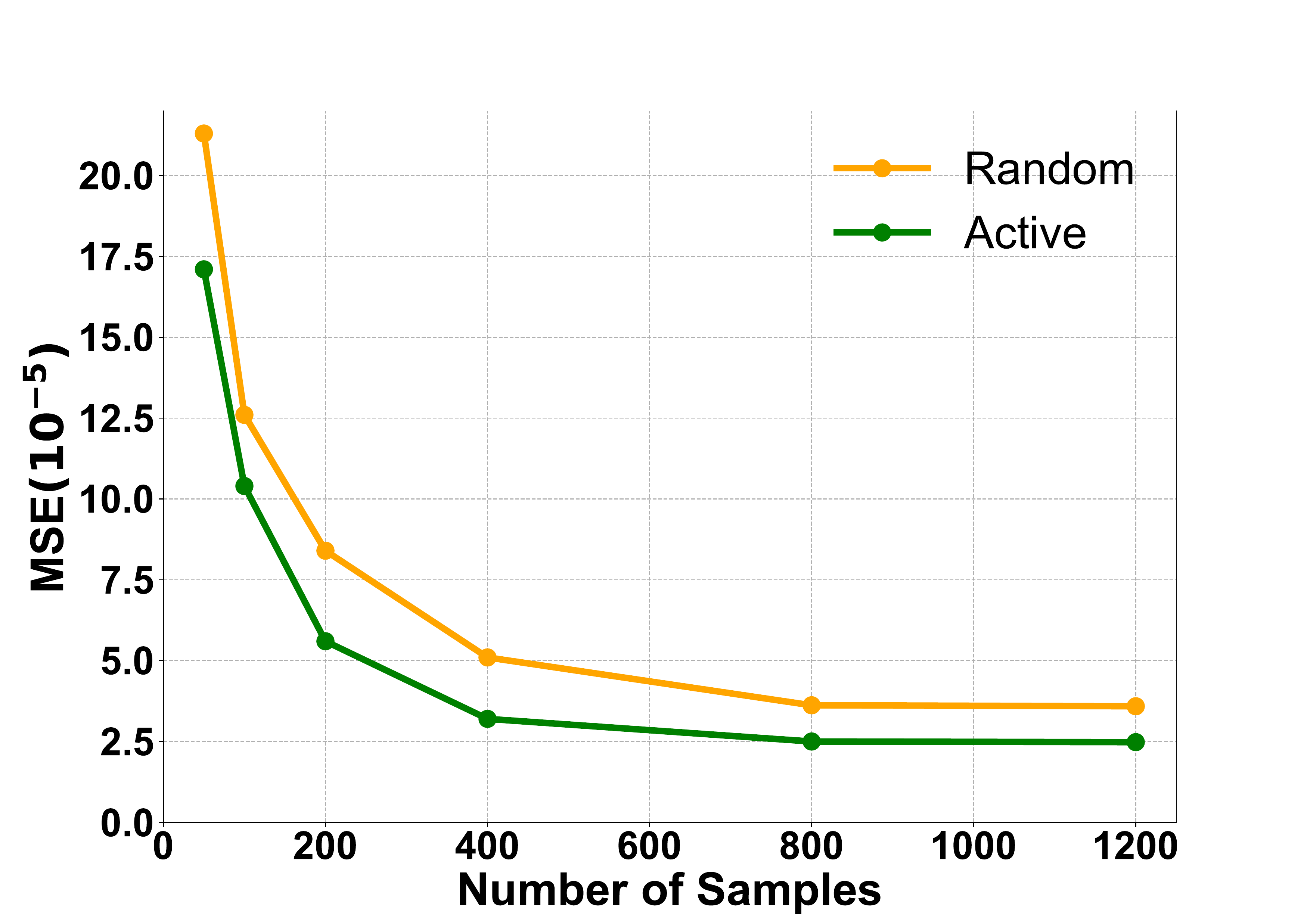}
			\end{minipage}
		}\\
		\vspace{-0.3cm}
		\subfigure[ResNet20]{
			\begin{minipage}[b]{0.44\linewidth}\label{fig:magnitude}
				\includegraphics[height=3.3cm, width=4.5cm]{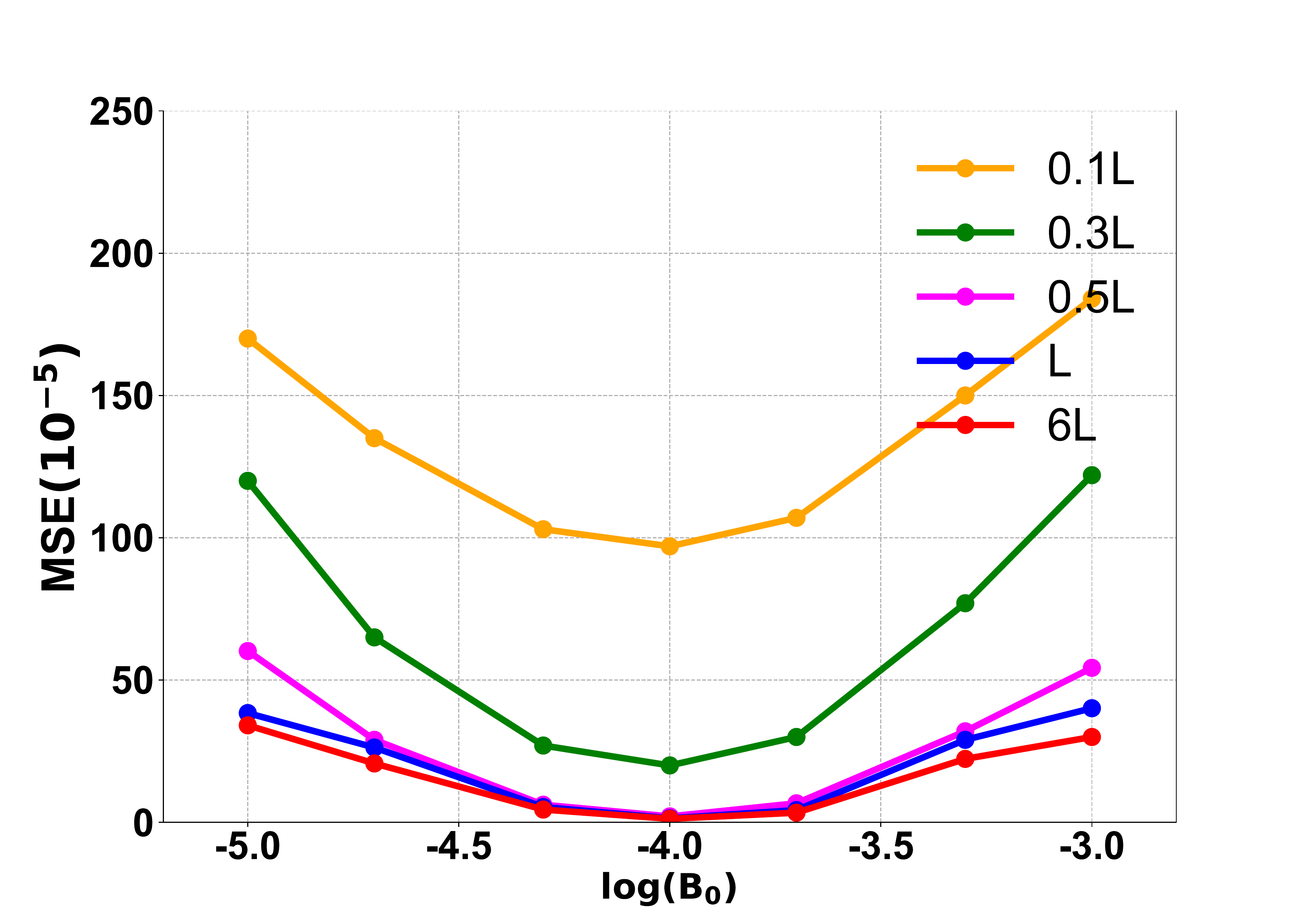}
			\end{minipage}
		}
		\hfill
		\subfigure[ResNet50]{
			\begin{minipage}[b]{0.47\linewidth}\label{fig:number}
				\includegraphics[height=3.3cm, width=4.5cm]{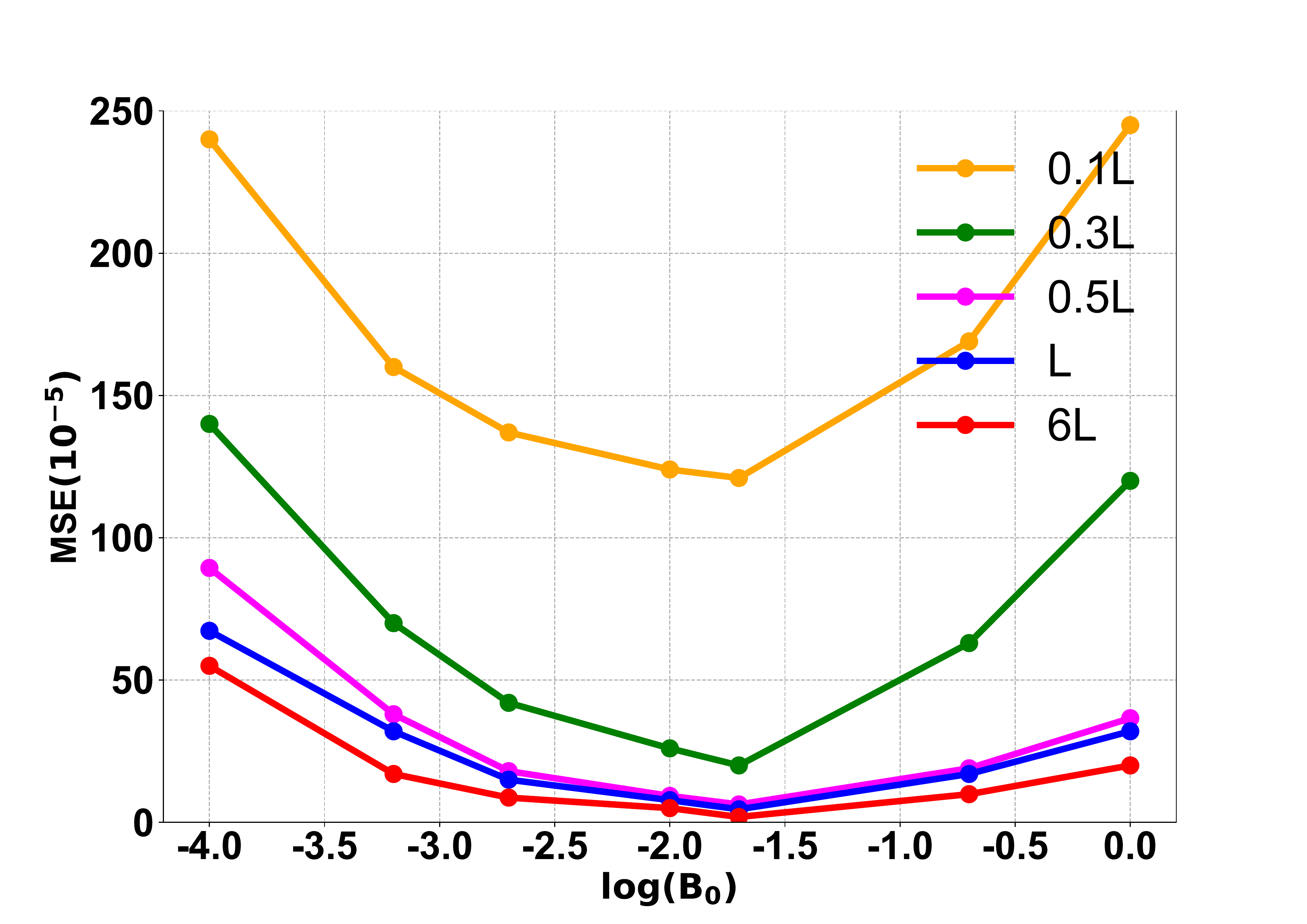}
			\end{minipage}
		}
		\caption{(a) and (b) report the MSE ($\times 10^{-5}$) between the predicted and actual accuracy of the sampled compression policies with different sampling strategies and varying numbers of samples for ResNet20 and ResNet50 architectures respectively. (c) and (d) show that with various perturbation magnitude and different numbers of sampled perturbed policies in uncertainty estimation for ResNet20 and ResNet50. }
		\label{MSE}
	\end{center}
\vspace{-0.2cm}
\end{figure}

\subsubsection{Effects of Performance Predictor Learning}
\textbf{Performance w.r.t. different sampling strategies and varying numbers of sampled lightweight models: }We trained the predictor with the actual accuracies of $50$, $100$, $200$, $400$, $800$ and $1,200$ compressed models obtained via random and active compression policy evaluation, where $50$ compression policies were also randomly sampled for validation. We depict the MSE between the actual and predicted accuracies in Figure \ref{MSE} (a) and (b) for ResNet20 and ResNet50. Our active sampling strategy chooses the uncertain compression policies that provide informative supervision for performance predictor learning, so that the predicted accuracy is more precise compared with the random sampling strategy. The advantages are more obvious for small training sets, which reveals the benefits of our active sampling for automated model compression in extremely low search cost. Training the performance predictor for ResNet50 requires more sampled policies to achieve low prediction error due to the larger search space. However, sampling more compression policies only slightly influences the MSE between the actual and predicted accuracies when the training set exceeds $800$ samples for the ResNet50 architecture, and we evaluated $800$ compression policies to learn the performance predictor in other experiments. The actual and predicted accuracies of randomly and actively sampled policies on different datasets across various network architectures are demonstrated in the appendix. 

\begin{table}
	\footnotesize
	\centering
	\caption{The accuracy on CIFAR-10, computational complexity and the search cost (GPU hours) of the optimal compression policy obtained by reinforcement learning, evolutionary algorithms and our compression policy optimization (CPO) under the computational cost constraint BOPs less than $0.2$G for ResNet20, where the reward for agents in reinforcement learning and fitness for population in evolutionary algorithms based on accuracy and model complexity were obtained via the learned performance predictor. The training cost is $0.58$ GPU hours for the lightweight models.}
	
	\label{Search_strategy}
	\vspace{0.1cm}
	\begin{tabular}{c|ccc}
		\hline
		& Acc.(\%) & BOPs(G) & Cost\\
		\hline
		Reinforcement learning & $92.03$ & $0.19$ & $0.34$\\
		Evolutionary algorithms & $92.10$ & $0.19$ & $0.50$\\
		CPO & $92.51$ & $0.20$ & $0.003$ \\
		\hline	
	\end{tabular}
\end{table}

\textbf{Impacts of perturbation magnitudes and the number of sampled perturbed policies in uncertainty estimation: }The hyperparameter $B_0$ represents the model capacity changes caused by perturbed compression policies, and we employed different settings for $B_0$ and show the influence on performance predictor learning. Given the perturbation magnitude of model capacity, we randomly sampled various numbers of perturbed compression policies and report the MSE between predicted and actual accuracies. $50$ randomly sampled compression policies were utilized for validation. Figure \ref{MSE} (c) and (d) demonstrate the results on ResNet20 and ResNet50 respectively. Medium $B_0$ results in the minimal MSE in both architectures, as small perturbation fails to collect sufficient information for uncertainty estimation and large one considers non-local information that has little contribution to uncertainty. Meanwhile, the optimal perturbation magnitude in ResNet50 is larger than ResNet20 due to the higher original model complexity of backbone networks. Sampling more perturbed compression policies positively contributes to the precision of the performance predictor learning because of more accurate uncertainty estimation. However, sampling over $0.5L$ perturbation for each policy only slightly improves the uncertainty estimation, while the computational cost increases significantly. To maintain high computational efficiency, we randomly sampled $0.5L$ perturbation of each policy for uncertainty estimation in the rest experiments.

\begin{table*}
	\scriptsize
	\centering
	\caption{The BOPs(G), compression ratio and the accuracy on CIFAR-10 of the obtained lightweight ResNet20 architectures. The existence of barrier complexity loss (Bar.) and the rounded gap minimization (Gap.) in the learning objective was varied. F$\backslash$M, F\&M, A$\backslash$M and A\&M respectively stand for fixed stepsizes without momentum, fixed stepsizes with momentum, adaptive stepsizes without momentum and adaptive stepsizes with momentum.}
	
	\label{ablation_objective}
	\renewcommand\arraystretch{1.2}
	\begin{tabular}{p{0.8cm}<{\centering}|p{0.8cm}<{\centering}|p{0.8cm}<{\centering}p{0.8cm}<{\centering}p{0.8cm}<{\centering}|p{0.8cm}<{\centering}p{0.8cm}<{\centering}p{0.8cm}<{\centering}|p{0.8cm}<{\centering}p{0.8cm}<{\centering}p{0.8cm}<{\centering}|p{0.8cm}<{\centering}p{0.8cm}<{\centering}p{0.8cm}<{\centering}}
		\hline
		\multirow{2}{*}{Bar.}&\multirow{2}{*}{Gap.}&\multicolumn{3}{c|}{F$\backslash$M}&\multicolumn{3}{c|}{F\&M}&\multicolumn{3}{c|}{A$\backslash$M}&\multicolumn{3}{c}{A\&M}\\
		\cline{3-14}
		&& BOPs & Comp. & Top-1& BOPs & Comp. & Top-1& BOPs & Comp.& Top-1& BOPs & Comp. & Top-1\\
		\hline
		\multirow{2}{*}{$\times$}& $\times$& $0.193$ & $216.58$ & $88.41$& $0.188$ & $222.34$ & $88.58$& $0.197$ & $212.43$ & $90.39$& $0.196$ & $213.27$ & $90.63$\\
		\cline{2-14}
		&$\checkmark$& $0.183$ & $227.47$ & $89.10$& $0.192$ & $217.58$ & $89.25$& $0.196$ & $213.27$ & $91.19$& $0.198$ & $211.11$ & $91.38$\\
		\hline
		\multirow{2}{*}{$\checkmark$}& $\times$ & $0.195$ & $214.35$ & $88.23$& $0.193$ & $216.58$ & $88.79$& $0.191$ & $218.85$ & $90.94$& $0.192$ & $217.71$ & $91.22$ \\
		\cline{2-14}
		& $\checkmark$ & $0.198$ & $211.11$ & $89.35$& $0.197$ & $212.43$ & $89.64$& $0.200$ & $209.00$ & $91.75$& $0.200$ & $209.00$ & $\bm{92.51}$ \\
		\hline	
	\end{tabular}
\vspace{-0.5cm}
\end{table*}

\subsubsection{Effects of Compression Policy Optimization}
\textbf{Comparison with other search strategies: }To validate the effectiveness and the efficiency of our ultrafast compression policy optimization, we compare the accuracy and the computational complexity of the optimal lightweight models searched by reinforcement learning and evolutionary algorithms, where the reward for agents in reinforcement learning and the fitness for population in evolutionary algorithms were obtained via the learned performance predictor. For reinforcement learning, we modified the implementations in \citep{wang2019haq} by adding the pruning ratio in the state and action space. For evolutionary algorithms, we leveraged the pipeline in \citep{wang2020apq} where the branch of architecture search was removed. The detailed implementations of reinforcement learning and evolutionary algorithms are demonstrated in the appendix. Table \ref{Search_strategy} demonstrates the accuracy, model complexity and the search cost of different search algorithms. Our ultrafast compression policy optimization acquires highest accuracy within the computational complexity constraint, and the search cost can be negligible compared with reinforcement learning and evolutionary algorithms.

\textbf{Performance w.r.t. different terms in learning objectives and various techniques in policy update: }Table \ref{ablation_objective} demonstrates the BOPs, compression ratio and accuracy of obtained light-weight networks with various objectives and update techniques. The existence of barrier complexity loss and the rounded gap minimization in the learning objectives was varied. The impacts of the presented adaptive stepsizes and the momentum-based gradient in compression policy update were also investigated. Comparing the model complexity and the accuracies across different rows, we conclude that more computational resource under the constraint is utilized with accuracy improvement via the barrier complexity loss. Meanwhile, the rounded policy gap minimization shrinks the difference between the optimal policy and the discrete one for deployment, and better accuracy-complexity trade-off is achieved. Comparing the performance across various columns, we observe that the adaptive stepsizes yield lightweight models with better performance within the expected complexity because of stable optimization process. Moreover, the gradient momentum provides historical information of the optimization process so that the obtained compression policy can escape the local maximum.

\begin{figure}[t]
	\centering
	\includegraphics[height=3.2cm, width=8.5cm]{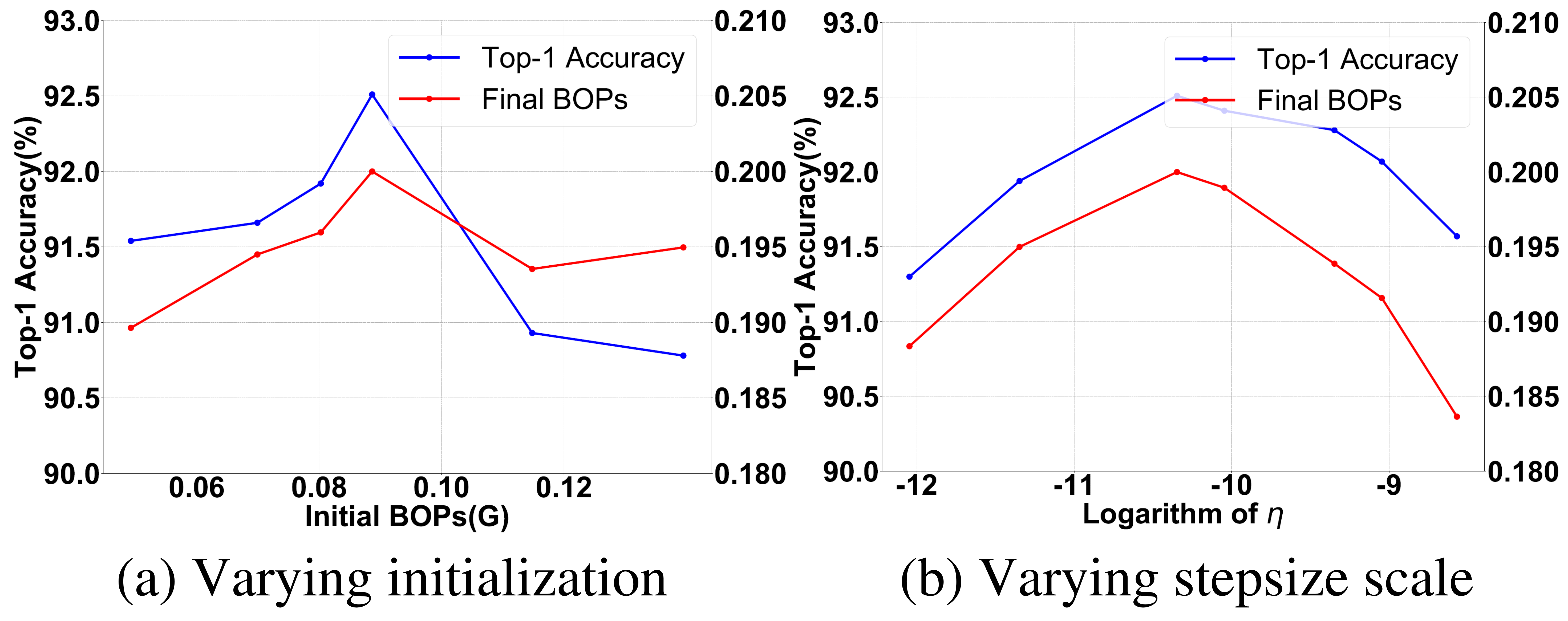}
	\vspace{-0.3cm}
	\caption{The actual accuracy and computation complexity of the optimal lightweight models obtained by our compression policy optimization with (a) different initialization and (b) various update stepsize scale under the computational cost constraint BOPs less than $0.2$G, where the ResNet20 architecture on CIFAR-10 was evaluated.}
	\vspace{-0cm}    
	\label{init_step}
\end{figure}

\textbf{Performance w.r.t. different compression policy initialization: }To investigate the influence of the compression strategy initialization on the performance of our compression policy optimization, we show the actual accuracies and the computation complexity of the optimal lightweight models w.r.t. the complexity of initialized compression policy in Figure \ref{init_step} (a). The results show that medium complexity for initialized compressed models acquires the highest accuracy given the complexity constraint. High complexity for initialized lightweight models attains local minimum during the compression policy update, while low complexity for initialization cannot converge to the optimal compressed models before reaching the complexity constraint.

\iffalse
\begin{figure}
	\centering
	\vspace{-0.2cm}
	\begin{center}
		\subfigure[Varying initialization]{
			\hspace{-0.6cm}
			\begin{minipage}[b]{0.44\linewidth}\label{fig:Initialization}
				\includegraphics[height=3.1cm, width=4cm]{pic/initialization.pdf}
			\end{minipage}
		}
		\hfill
		\subfigure[Varying stepsize scale]{
			\begin{minipage}[b]{0.48\linewidth}\label{fig:Stepsizes}
				\hspace{-0.3cm}
				\includegraphics[height=3.1cm, width=4cm]{pic/stepsize.pdf}
			\end{minipage}
		}
		\caption{The actual accuracy and computation complexity of the optimal lightweight models obtained by our compression policy optimization with (a) different initialization and (b) various update stepsize scale under the computational cost constraint BOPs less than $0.2$G, where the ResNet20 architecture on CIFAR-10 was evaluated.}
	\end{center}
\vspace{-0.2cm}
\end{figure}
\fi

\textbf{Performance w.r.t. various update stepsize scale $\eta$: }The update stepsize scale is controlled by the hyperparameter $\eta$ in (\ref{ada_stepsize}), where low $\eta$ generally leads to small stepsizes and vice versa. Figure \ref{init_step} (b) depicts the actual accuracy and the computational complexity of the optimal compressed models acquired via different parameter settings of $\eta$. Medium stepsizes outperform other choices. Small stepsizes fail to achieve the optimal compression policy when reaching the maximum update iterations due to the local maximum, and large stepsizes enforce ultrafast compression policy optimization to be hard to converge.

\begin{table}
	\renewcommand\arraystretch{1.15}
	\scriptsize
	\centering
	\caption{Comparison of network complexity, classification accuracy and search cost on CIFAR-10 with state-of-the-art network compression methods in VGG-small and ResNet20. W/A means the bitwidth of weights and activations respectively, and the computational complexity is measured by MACs (G) and BOPs (G). Comp. represents the network compression ratio calculated by the BOPs, and Acc. means the accuracy on image classification. The search cost is demonstrated by GPU hours, where $N$ means the number of deployment scenarios. The number in the bracket of search cost for mixed-precision quantization demonstrates the break-even point of baseline methods whose search cost is higher than our SeerNet. The training cost for each model of VGG-small and ResNet20 is $0.43$ and $0.58$ GPU hours respectively.}
	\label{CIFAR-10}
	\begin{tabular}{p{1.1cm}<{\centering}|p{0.6cm}<{\centering}p{0.6cm}<{\centering}p{0.6cm}<{\centering}p{0.6cm}<{\centering}p{0.6cm}<{\centering}|p{1.2cm}<{\centering}}	
		\hline
		Methods & W/A &  MACs & BOPs & Comp. & Acc. & Cost\\
		\hline\hline
		\multicolumn{7}{c}{VGG-small}\\
		\hline
		Baseline & $32/32$&$0.494$&$506.0$&$-$& $92.80$ &$-$\\
		\hline	
		LQ-Nets & $4/4$ & $0.494$&$7.91$&$64.00$& $92.12$ & $-$\\
		ALQ & mixed & $0.494$&$7.44$&$68.00$& $90.90$ & $~~1.3N (11)$\\
		SeerNet& mixed & $0.459$ & $7.35$ &$68.84$ & $92.97$ &$13$+$0.003N$ \\
		\hline
		DQ$^{*}$& mixed & $0.613$ & $3.40$ & $185.00$ & $91.59$ &$~~0.9N (15)$\\
		SeerNet& mixed & $0.340$ & $3.88$ & $130.41$ & $92.84$ & $13$+$0.003N$\\
		\hline
		
		DJPQ$^{*}$&mixed & $0.367$ & $2.99$ & $210.37$ & $ 91.54 $ &$~~0.5N (27)$\\
		SeerNet& mixed & $0.223$ & $2.27$ & $222.91$ & $92.69$ & $13$+$0.003N$\\
		
		\hline\hline
		\multicolumn{7}{c}{ResNet20}\\
		\hline
		Baseline& $32/32$ &$0.041$&$41.8$&$-$&$92.51$&$-$\\
		\hline
		%LQ-Nets & $2/32$ & $0.041$&$2.62$&$16.00$& $92.40$ & $-$\\
		APoT & $4/4$ & $0.041$&$0.65$&$64.00$& $92.45$ & $-$\\
		HMQ& mixed &$0.041$& $0.65$& $64.00$ & $92.59$ & $~~2.0N (11)$\\
		SeerNet& mixed &$0.038$ & $0.63$ & $65.02$ & $92.71$ &$21$+$0.003N$\\
		\hline
		BP-NAS& mixed & $0.041$& $0.39$& $106.35$ & $92.12$ & $~~1.1N (20)$\\
		SeerNet& mixed &$0.030$ & $0.34$ & $122.94$ & $92.55$ &$21$+$0.003N$\\
		\hline
		BP-NAS& mixed & $0.041$& $0.32$& $128.97$ & $92.04$ & $~~1.1N (20)$\\
		SeerNet& mixed &$0.038$ & $0.20$ & $209.00$ & $92.51$&$21$+$0.003N$\\
		\hline		
	\end{tabular}
	\vspace{0.2cm}
\end{table}

\iffalse
We also show the performance of the two-stage compression consisting of independent pruning and quantization, where the automated pruning method VIB \citep{dai2018compressing} was applied to combine with DQ.
\fi

\subsection{Comparison with the State-of-the-art Methods}
We compare our SeerNet with the fixed-precision quantization methods including LQ-Nets \citep{zhang2018lq}, APoT \citep{li2019additive}, RQ \citep{louizos2018relaxed}, LSQ \citep{esser2019learned}, AdaBits \citep{jin2020adabits}, BitMixer \citep{bulat2021bit} and mixed-precision quantization approaches such as ALQ \citep{qu2020adaptive}, DQ \citep{uhlich2019differentiable}, BP-NAS \citep{yu2020search}, HAQ \citep{wang2019haq}, HMQ \citep{habi2020hmq}, HAWQ \citep{dong2019hawq}.  Meanwhile, we compare the accuracy with the state-of-the-art automated model compression method DJPQ \citep{wang2020differentiable} where the pruning and quantization policies were jointly searched. In order to show the performance in different accuracy-complexity trade-offs, we leveraged three BOPs constraints for each architecture. The reduction in MACs demonstrates the network pruning ratio, and the BOPs decrease reveals the total compression effect. Therefore, we define the reduction ratio of BOPs as the compression ratio. The reported search cost only contains computational cost to obtain the optimal compression policy, and that of baseline methods is evaluated by rerunning the released code or our re-implementation. The total cost for model deployment can be easily calculated by summing the search cost and the training cost. The break-even points indicates the number of scenarios where the search cost is higher than our SeerNet. The acquired compression policy of our SeerNet for different architectures is visualized in our appendix. 

\begin{table}[t]
	\renewcommand\arraystretch{1.1}
	\scriptsize
	\centering
	\caption{The BOPs(G), top-1 classification accuracy and search cost on ImageNet with state-of-the-art network compression methods in MobileNet-V2, ResNet18 and ResNet50. The training cost for each model of MobileNet-V2, ResNet18 and ResNet50 is $37.4$, $60.8$ and $80.9$ GPU hours respectively.}
	\vspace{0cm}              
	\label{ImageNet}
	\begin{tabular}{p{1.1cm}<{\centering}|p{0.6cm}<{\centering}p{0.55cm}<{\centering}p{0.55cm}<{\centering}p{0.55cm}<{\centering}p{0.75cm}<{\centering}|p{1.3cm}<{\centering}}	
		\hline
		Methods & W/A & MACs & BOPs & Comp. & Top-1 & Cost\\
		\hline\hline
		\multicolumn{7}{c}{MobileNet-V2}\\
		\hline
		Baseline & $32/32$ & $0.33$ & $337.9$ &$-$& $71.72$ &$-$\\
		\hline
		RQ & $6/6$ &$0.33$ &$11.88$&$28.44$&$68.02$& $-$\\
		HMQ & mixed & $0.33$ & $10.97$ & $30.80$ & $71.40$ & $~~31.4N (24)$\\
		SeerNet & mixed & $0.25$ & $10.82$  & $31.22$ & $71.47$ & $750$+$0.004N$ \\
		
		\hline
		HAQ &  mixed &$0.33$ &$8.25$ & $40.96$&$69.45$&$~~51.1N (15)$\\
		DJPQ & mixed & $0.28$ &$7.87$&$42.96$&$69.30$ &$~~12.2N (62)$\\
		SeerNet & mixed & $0.22$ & $7.69$  & $43.91$ & $70.76$ & $750$+$0.006N$ \\
		\hline
		HMQ& mixed &$0.33$ &$5.25$& $64.40$&$70.90$&$~~33.5N (23)$\\
		DQ &  mixed & $0.33$&$4.92$&$68.67$&$69.74$& $~~21.6N (35)$\\
		SeerNet & mixed & $0.19$ & $4.88$  & $69.26$ & $70.55$ & $750$+$0.006N$ \\

		\hline\hline
		\multicolumn{7}{c}{ResNet18}\\
		\hline
		Baseline& $32/32$ &$1.81$ & $1853.4$& $-$ & $69.74$ &$-$\\
		\hline
		ALQ &mixed &$1.81$& $58.50$& $31.68$& $67.70$  & $~~34.7N (15)$\\
		SeerNet& mixed & $1.37$ &$56.94$ & $32.55$ & $69.72$ & $500$+$0.003N$\\
		\hline
		DJPQ & mixed &$1.39$&$35.01$&$52.94$& $69.27$ &$~~18.2N (28)$\\
		HAWQ & mixed & $1.81$&$34.00$&$54.51$&$68.45$&$~~22.7N(23)$\\	
		SeerNet& mixed & $1.22$ &$31.84$ & $58.21$ & $69.48$ & $500$+$0.004N$\\
		\hline
		LSQ & $3/2$ &$1.81$ &$10.86$& $170.66$&$66.90$&$-$\\
		BitMixer & $2/2$ &$1.81$ &$7.24$& $256.00$&$64.40$&$-$\\
		ALQ& mixed &$1.81$ &$7.24$& $256.00$&$66.40$&$~~38.5N (13)$\\
		SeerNet& mixed & $0.70$ &$7.19$ & $257.71$ & $67.84$ & $500$+$0.004N$\\
		\hline\hline
		\multicolumn{7}{c}{ResNet50}\\
		\hline
		Baseline& $32/32$ &$3.86$ & $3952.6$& $-$ & $76.40$ &$-$\\
		\hline
		HAQ & mixed &$3.86$& $94.92$& $41.64$& $75.30$  & $~~67.2N (15)$\\
		SeerNet& mixed & $3.34$ &$91.99$ & $42.97$ & $76.70$ & $950$+$0.005N$\\
		\hline
		HAWQ & mixed &$3.86$&$61.29$&$64.49$ &$75.48$&$~~34.5N (28)$\\
		LQ-Net & $4/4$ &$3.86$&$61.76$&$64.00$ &$75.10$&$-$\\
		AdaBits & $4/4$ &$3.86$&$61.76$&$64.00$ &$76.10$&$-$\\
		SeerNet& mixed & $2.97$ &$59.66$ & $66.25$ & $76.61$ & $950$+$0.006N$\\
		\hline
		AdaBits & $3/3$ &$3.86$&$34.74$&$113.78$ &$75.80$&$-$\\
		HMQ& mixed &$3.86$ &$37.72$& $104.8$&$75.45$&$~~49.4N (20)$\\
		BP-NAS& mixed & $3.86$&$33.22$ &$118.98$ & $75.71$ & $~~35.6N (27)$ \\
		SeerNet& mixed & $2.12$ &$31.67$ & $124.81$ & $75.90$ & $950$+$0.007N$\\
		\hline
	\end{tabular}
	%\vspace{0.35cm}
	\vspace{0.05cm}
\end{table}

\subsubsection{Comparison on Image Classification}
\vspace{-0.2cm}
\textbf{Comparision on CIFAR-10: }Table \ref{CIFAR-10} shows the experimental results on CIFAR-10 with VGG-small and ResNet20. $*$ represents the methods utilizing the VGG7 architecture which is very similar to VGG-small. The fixed-precision quantization ignores the importance variety among different layers and fails to effectively assign the optimal bitwidth for each layer. The mixed-precision quantization does not consider the redundancy in different channels while the pruning strategy can further enhance the model efficiency. Compared with the state-of-the-art mixed-precision networks BP-NAS, our SeerNet enhances the accuracy by $0.47$\% ($92.51$\% vs. $92.04$\%) with $1.60\times$ BOPs reduction ($0.20$G vs. $0.32$G) with the ResNet20 architecture. Although the automatic model compression method DJPQ jointly searches the pruning and quantization policy, the search deficiency leads to heavy computational cost due to the extremely large search space. On the contrary, our SeerNet obtains the optimal lightweight model without complex compression policy search and evaluation, so that the proposed method only requires $0.003$ GPU hour to marginally search an automated model compression policy on both VGG-small and ResNet20. Because the number of deployment scenarios is usually very large in realistic applications with frequent changes of hardware configurations and battery levels, the compression policy search cost is reduced sizably in our SeerNet.

\textbf{Comparison on ImageNet: }The results on ImageNet with MobileNetV2, ResNet18 and ResNet50 are demonstrated in Table \ref{ImageNet}. Mixed-precision quantization outperforms fixed-precision quantization by a larger margin on ImageNet compared with CIFAR-10 as the optimal bitwidth assignment makes more contribution to the image classification on challenging datasets. Compared with the state-of-the-art mixed-precision quantization method BP-NAS, our SeerNet improves the top-1 accuracy by $0.19\%$ ($75.90$\% vs. $75.71$\%) and decreases BOPs by $1.05\times$ ($31.67$G vs. $33.22$G) with the ResNet50 architecture due to the automated pruning strategy. Meanwhile, the marginal search cost is decreased by $5086\times$ ($35.6$ GPU hours vs. $0.007$ GPU hours). Compared with the state-of-the-art automatic model compression method, our SeerNet directly optimizes the compression policy with the discriminative performance predictor without resource-exhaustive compression policy search and evaluation process. We obtain better accuracy-complexity trade-off with only $0.05\%$ and $0.02\%$ marginal search cost compared with DJPQ in MobileNet-V2 and ResNet18 respectively.  The marginal cost reduction is much more sizable on ImageNet compared with CIFAR-10 due to the significantly increased training cost for each sampled compression policy in exhaustive evaluation, which shows the superiority of our SeerNet in deployment facing largescale datasets.

\begin{table}
	\renewcommand\arraystretch{1.1}
	\scriptsize
	\centering
	\caption{Comparison of BOPs(G), mean average precision and search cost on PASCAL VOC with existing network compression methods, where the SSD framework with VGG16 and Faster R-CNN with ResNet18 was employed. The training cost for each model of VGG16 and ResNet18 is $19.5$ and $18.7$ GPU hours respectively.}
	\label{VOC}
	\begin{tabular}{p{1.1cm}<{\centering}|p{0.6cm}<{\centering}p{0.6cm}<{\centering}p{0.6cm}<{\centering}p{0.6cm}<{\centering}p{0.6cm}<{\centering}|p{1.3cm}<{\centering}}	
		\hline
		Methods & W/A &  MACs & BOPs & Comp. & mAP & Cost\\
		\hline\hline
		\multicolumn{7}{c}{SSD \& VGG16}\\
		\hline
		Baseline & $32/32$&$27.14$&$27787.7$&$-$& $72.4$ &$-$\\
		\hline	
		%		APoT & $2/32$ & $0.613$&$31.63$&$16.00$& $91.80$ & $-$\\
		HAQ & mixed & $27.14$&$846.67$&$32.82$& $68.9$ & $~~48.8N (17)$\\
		SeerNet& mixed & $18.23$ & $768.25$ &$36.17$ & $69.7$ &$800$+$0.005N$ \\
		\hline
		DJPQ & mixed & $16.11$&$627.12$&$44.31$& $66.8$ & $~~29.1N (28)$\\
		SeerNet& mixed & $14.68$ & $622.76$ &$44.62$ & $68.8$ &$800$+$0.005N$ \\
		\hline
		
		BP-NAS&mixed & $27.14$ & $453.60$ & $61.26$ & $ 66.7 $ &$~~33.2N (25)$\\
		SeerNet& mixed & $13.59$ & $435.27$ &$63.84$ & $67.3$ &$800$+$0.006N$ \\
		
		\hline\hline
		\multicolumn{7}{c}{Faster R-CNN \& ResNet18}\\
		\hline
		Baseline& $32/32$ &$22.01$&$22534.8$&$-$&$74.5$&$-$\\
		\hline
		%LQ-Nets & $2/32$ & $0.041$&$2.62$&$16.00$& $92.40$ & $-$\\
		APoT& $5/5$ & $22.01$& $550.17$& $40.96$ & $71.2$ & $-$\\
		SeerNet& mixed &$19.04$ & $526.39$ & $42.81$ & $73.4$&$650$+$0.005N$\\
		\hline
		HMQ & mixed & $22.01$&$343.41$&$65.62$& $71.3$ & $~~28.9N (23)$\\
		DJPQ& mixed &$17.19$& $349.54$& $64.47$ & $69.7$ & $~~27.2N (24)$\\
		SeerNet& mixed &$15.21$ & $339.64$ & $66.35$ & $72.3$ &$650$+$0.005N$\\
		\hline
		BP-NAS& mixed & $22.01$& $298.63$& $75.46$ & $69.7$ & $~~22.4N (30)$\\
		SeerNet& mixed &$13.22$ & $298.43$ & $75.51$ & $71.8$ &$650$+$0.006N$\\
		\hline
	\end{tabular}
%\vspace{0.35cm}
\end{table}

\begin{table*}[t]
	\renewcommand\arraystretch{1.1}
	\scriptsize
	\centering
	\caption{Comparison of BOPs(G), mAP@$[.5, .95]$ and search cost on COCO with state-of-the-art network compression methods in the SSD framework with VGG16 and Faster R-CNN with ResNet18. The average precision at different IoU thresholds and that for objects in various sizes are also illustrated. The training cost of SSD and Faster R-CNN is $56.5$ and $53.2$ GPU hours.}
	\label{COCO}
	\begin{tabular}{p{1.1cm}<{\centering}|p{1cm}<{\centering}p{1cm}<{\centering}p{1cm}<{\centering}p{1cm}<{\centering}|p{0.9cm}<{\centering}p{0.9cm}<{\centering}p{0.9cm}<{\centering}p{0.9cm}<{\centering}p{0.9cm}<{\centering}p{0.9cm}<{\centering}|p{1.8cm}<{\centering}}	
		\hline
		Methods & W/A &  MACs & BOPs & Comp. & mAP & $AP_{50}$ & $AP_{75}$ & $AP_{s}$ & $AP_{m}$ & $AP_{l}$ & Cost\\
		\hline\hline
		\multicolumn{12}{c}{SSD \& VGG16}\\
		\hline
		Baseline& $32/32$ &$27.14$ & $27787.7$& $-$ & $23.2$ &$41.2$&$23.4$ & $5.3$& $23.2$ & $39.6$ &$-$\\
		\hline
		DJPQ& mixed & $18.39$ &$795.98$ & $34.91$ & $20.1$ & $37.9$&$20.6$& $5.3$& $22.3$& $34.7$  & $~53.6N (26)$\\
		SeerNet& mixed & $21.90$ &$783.86$ & $35.45$ & $22.7$ & $40.2$&$22.8$& $5.8$& $24.4$& $35.9$  & $~1350+0.005N$\\
		\hline
		BP-NAS & mixed & $27.14$& $616.14 $ & $45.10$& $20.8 $ & $38.4 $ & $ 20.7$&$5.2 $& $22.5 $& $ 33.8$& $~42.6N (32) $  \\
		SeerNet& mixed & $19.03$ &$569.89$ & $48.76$ & $22.1$ & $39.9$&$22.6$& $6.0$& $24.0$& $36.1$  & $~1350+0.005N$\\
		\hline
		
		HAQ& mixed & $ 27.14$ &$ 445.67$ & $ 62.35$ & $20.1 $ & $ 37.5$&$ 19.9$& $5.2 $& $21.3 $& $32.6 $  & $~95.3N (15)$\\
		APoT & 4/4 & $27.14$ &$434.18$ & $64.00$ & $18.1$ & $34.4$&$17.5$& $4.5$& $19.1$& $29.4$  & $-$\\
		SeerNet& mixed & $15.51$ &$421.73$ & $65.89$ & $21.3$ & $39.0$&$21.3$& $5.7$& $23.5$& $34.2$  & $~1350+0.006N$\\
		\hline\hline
		\multicolumn{12}{c}{Fatser R-CNN \& ResNet18}\\
		\hline
		Baseline& $32/32$ &$22.01$ & $22534.8$& $-$ & $26.0$ &$44.8$&$27.2$& $10.0$& $28.9$& $39.7$  & $-$\\
		\hline
		HAQ & mixed &$22.01$& $471.83$& $47.76$& $25.5$  & $44.0$&$26.3$& $12.8$& $27.5$& $33.8$  & $~89.9N (14)$\\
		
		SeerNet& mixed & $19.31$ &$458.93$ & $49.10$ & $26.8$ & $45.7$ &$28.1$& $13.8$& $29.3$& $35.0$  & $ 1200+0.005N $\\
		\hline
		DJPQ & mixed &$18.24$&$342.32$&$65.83$ &$24.4$&$40.4$&$25.6$& $11.4$& $25.6$& $30.7$  & $~47.3N (26)$\\
		APoT & $4/4$ &$22.01$&$352.11$&$64.00$ &$23.2$&$39.9$&$24.1$& $11.6$& $25.3$& $30.1$  & $-$\\
		SeerNet& mixed & $14.25$ &$327.59$ & $68.79$ & $25.5$ & $44.4$ &$26.3$& $12.6$& $27.9$& $33.8$  & $ 1200+0.005N $\\
		\hline
		HMQ& mixed &$22.01$ &$301.23$& $74.81$&$24.1$&$42.8$&$24.5$& $12.7$& $26.9$& $30.4$  & $~55.5N (22)$\\		
		BP-NAS& mixed & $22.01$&$312.38$ &$72.14$ & $23.6$ & $41.9$&$23.9$& $12.7$& $27.6$& $32.0$  & $~38.7N (32)$ \\
		SeerNet& mixed & $12.39$ &$282.60$ & $79.74$ & $25.1$ & $43.7$ &$25.9$& $13.8$& $28.2$& $32.9$  & $ 1200+0.006N $\\
		\hline
		
	\end{tabular}
	\vspace{-0.3cm}
\end{table*}

\subsubsection{Comparison on Object Detection}
\textbf{Comparison on PASCAL VOC: }Table \ref{VOC} illustrates the computational complexity and the mAP of different compression methods on PASCAL VOC, where the search cost does not contain the computational cost for the model pretraining on ImageNet. Similar to image classification, the mixed-precision networks achieve more optimal accuracy-complexity trade-offs with different computational cost constraints. Our SeerNet enhances the mAP of the state-of-the-art DJPQ by $2.6$\% ($72.3$\% vs. $69.7$\%) in the Faster R-CNN framework with VGG16, where the BOPs are similar. Moreover, SeerNet only requires $0.005$ GPU hour compared with $27.2$ GPU hours in DJPQ for marginal compression policy search. The significant search efficiency improvement enables flexible model deployment for different hardware configurations and battery levels in realistic applications relying on object detection such as autonomous driving \citep{chen2017learning}, which usually requires models with hundreds of complexity constraints.

\textbf{Comparison on COCO: }Despite of the BOPs and the mAP on COCO, we also show the average precision at different IoU thresholds and that for objects in various sizes. Table \ref{COCO} depicts the results, where our SeerNet achieves better accuracy-complexity trade-offs than conventional mixed-precision networks and automatic model compression methods across different detection frameworks and backbone architectures. SeerNet is free of complex policy search and evaluation stage, and we only require $0.02$\% search cost ($0.006$ GPU hours vs. $38.7$ GPU hours) to marginally acquire the promising pruning and quantization policy for the Faster R-CNN detector with ResNet18 backbone in flexible deployment. Since training deep neural networks on the largescale COCO dataset costs much more computational resources, our SeerNet saves the computational cost of optimal compression policy acquisition more sizably compared with that trained on PASVAL VOC.

\section{Conclusion}
In this paper, we have presented the ultrafast automated model compression framework for flexible network deployment. The proposed SeerNet learns the accurate performance predictor in acceptable training cost via active compression policy evaluation, where the most uncertain pruning and quantization strategies with informative supervision are selected by efficient evolutionary search. Then the gradient that maximizes the predicted performance under the barrier complexity constraint is leveraged to differentiably search the desirable compression policy, where adaptive update stepsizes with momentum are employed to strengthen the optimality of the acquired pruning and quantization strategies. Therefore, ultrafast automated model compression is achieved without resource-exhaustive compression policy search and evaluation. Extensive experiments on image classification and object detection demonstrate the superiority in efficiency and effectiveness of the proposed method. There are two interesting directions for the future work: (1) extending our SeerNet to other network architectures such as transformers and graph neural networks, (2) implementing the SeerNet method with hardware cost constraint such as latency and energy.

\section*{Acknowledgments}
This work was supported in part by the National Key Research and Development Program of China under Grant 2017YFA0700802, in part by the National Natural Science Foundation of China under Grant 62125603, and in part by a grant from the Beijing Academy of Artificial Intelligence (BAAI).

\section*{Compliance with Ethical Standards}
\textbf{Conflict of interest} The authors declare that they have no conflict of interest.

\noindent\textbf{Ethical approval} This article does not contain any studies with human participants or animals.

\section*{Appendix}

\begin{figure*}[t]
	\centering
	\includegraphics[height=5.3cm, width=17.5cm]{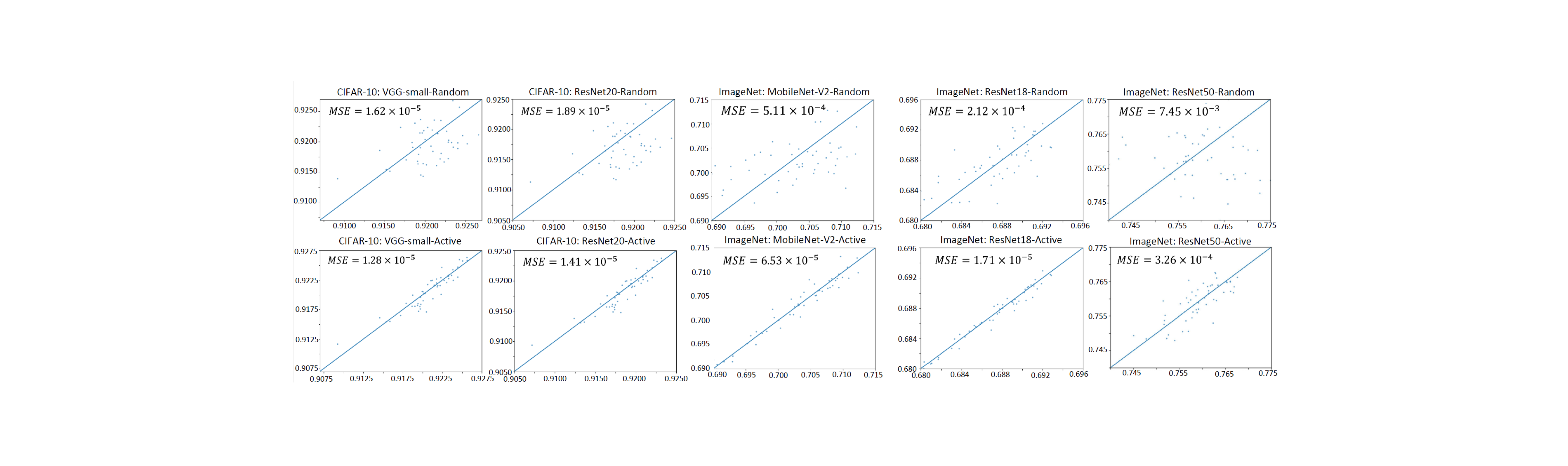}
	\caption{The actual and predicted accuracy for different compression policy across various datasets and architectures, where random sampling and our active sampling are both evaluated. The architectures for evaluation on CIFAR-10 include VGG-small and ResNet20, and those on ImageNet contain MobileNet-V2, ResNet18 and ResNet50. The horizontal axis represents the predicted accuracy and the vertical axis means the actual accuracy. The mean squared errors (MSE) are also demonstrated in the figure.}
	\vspace{-0.1cm}    
	\label{scattering}
\end{figure*}

\section*{A. Mathematical formulation from (11) to (12) of the manuscript.}
Since we leverage the deterministic neural networks to predict the accuracy of various lightweight models, we present the alternative objective function to tractably calculate the objective (11) in the manuscript based on importance sampling. We first rewrite the importance weight in (11) of the manuscript via the analytical form of Dirac-delta function:
\begin{align}
	\qquad\qquad\qquad \frac{p(a|\bm{s})}{p(a|\bm{\hat{s}})}=\frac{\lim\limits_{\epsilon\rightarrow 0}\frac{1}{\pi}\frac{\epsilon}{\epsilon^2+(a-a_{10})^2}}{\lim\limits_{\epsilon\rightarrow 0}\frac{1}{\pi}\frac{\epsilon}{\epsilon^2+(a-a_{20})^2}}
\end{align}where $a_{10}$ and $a_{20}$ are distribution parameters of original and perturbed policies parameterized by the performance predictor. As $\epsilon$ is higher-order infinitesimal of $a-a_{10}$ and $a-a_{20}$ according to the definition of Dirac-delta function, the importance weight can be rewritten as follows:
\begin{align*}
	&\frac{p(a|\bm{s})}{p(a|\bm{\hat{s}})}=\frac{(a-a_{20})^2}{(a-a_{10})^2}\notag\\
	&=\frac{(a-a_{10})^2+(a_{10}-a_{20})^2+2(a-a_{10})(a_{10}-a_{20})}{(a-a_{10})^2}
\end{align*}Since $a$ means the predicted accuracy parameterized by $a_{10}$, the difference between $a$ and $a_{10}$ can be assumed as a small constant $\gamma_{10}$ in the deterministic settings. Therefore, the difference between $a$ and $a_{10}$ is far less than that between $a_{10}$ and $a_{20}$. We obtain the following relationship between $\frac{p(a|\bm{s})}{p(a|\bm{\hat{s}})}$ and $a_{10}-a_{20}$:
\begin{align}
	\qquad~~ \frac{p(a|\bm{s})}{p(a|\bm{\hat{s}})}=(a_{10}-a_{20})^2/\gamma_{10} \propto (a_{10}-a_{20})^2
\end{align}$a_{10}$ and $a_{20}$ are represented by $f(\bm{s})$ and $f(\hat{\bm{s}})$, and the loss function $l(f(\bm{s}),a)$ of accuracy prediction is assigned with the $L_2$ norm of the difference between the predicted and actual accuracy $(f(\bm{s}^i)-a^i)^2$. Therefore, we optimize the alternative objective (12) in the manuscript to provide informative supervision for performance predictor learning.

\begin{figure*}[t]
	\centering
	\includegraphics[height=17cm, width=17.5cm]{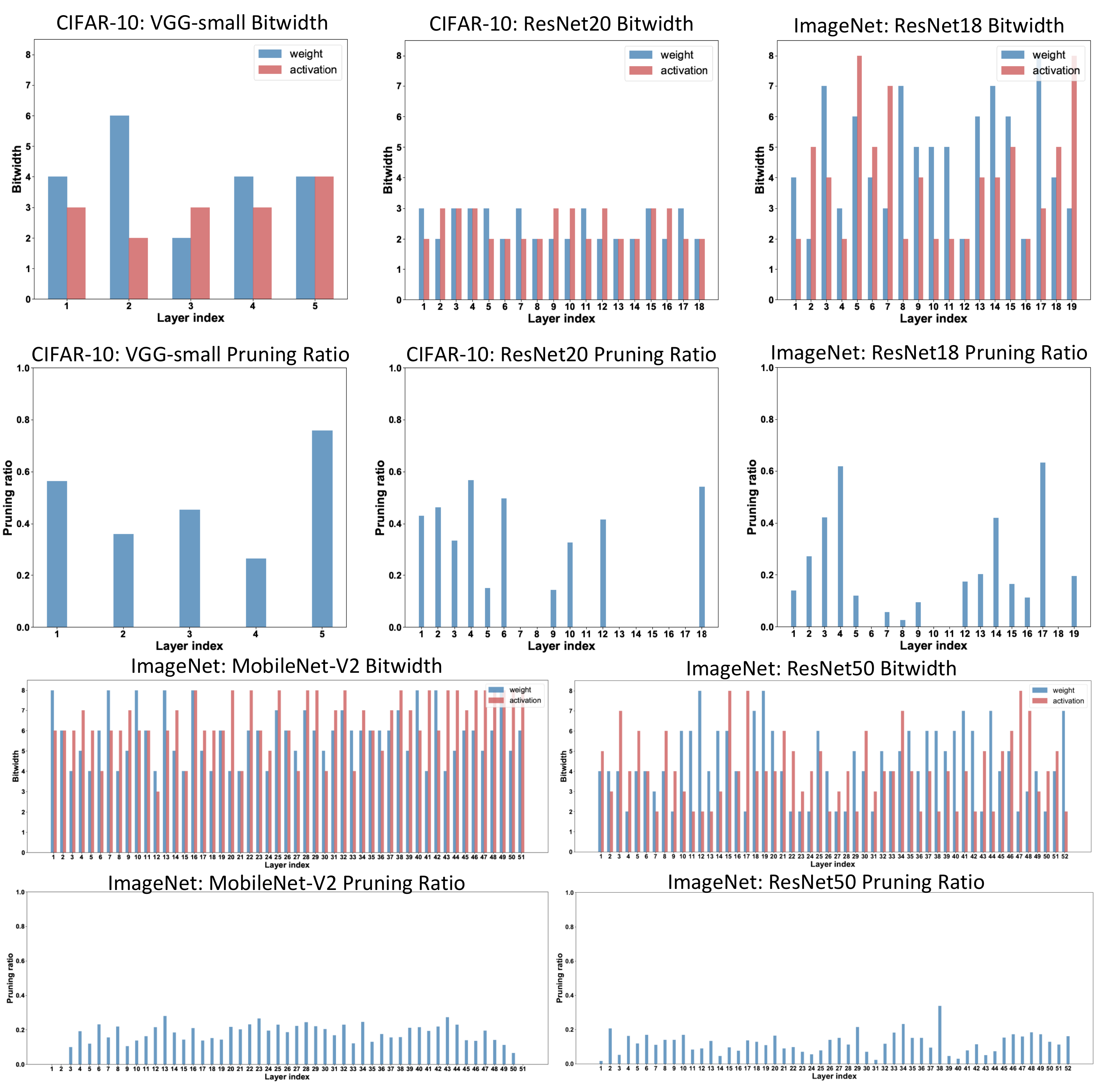}
	\caption{The visualization of the optimal compression policies obtained by our ultrafast SeerNet with given computational cost constraint on image classification. We utilized VGG-small and ResNet20 architectures on CIFAR-10 and MobileNet-V2,  ResNet18 and ResNet50 networks on ImageNet.}  
	\label{visualization}
	\vspace{-0cm}
\end{figure*}

\begin{table}
	\renewcommand\arraystretch{1.15}
	\footnotesize
	\centering
	\caption{The accuracy(\%) and BOPs(G) variance for Table 1 in the manuscript by running experiments for $5$ times.}
	
	\label{Search_strategy_variance}
	\vspace{0.1cm}
	\begin{tabular}{c|cc}
		\hline
		& Acc.(\%) & BOPs(G)\\
		\hline
		Reinforcement learning & $91.95\pm0.21$ & $0.19\pm0.01$ \\
		Evolutionary algorithms & $92.08\pm0.05$ & $0.19\pm0.01$ \\
		CPO & $92.38\pm0.17$ & $0.20\pm0$ \\
		\hline	
	\end{tabular}
\vspace{0.3cm}
\end{table}

\section*{B. The Actual and Predicted Accuracies for Sampled Lightweight Networks}
We employed the architectures of VGG-small \citep{zhang2018lq} and ResNet20 \citep{he2016deep} for automated model compression on CIFAR-10, and compressed MobileNet-V2 \citep{sandler2018mobilenetv2}, ResNet18 and ResNet50 architectures on ImageNet. We trained the performance predictor by the accuracy of 800 randomly sampled and 800 actively sampled compressed models respectively, and regressed the accuracy for 50 randomly sampled lightweight models via the well-trained performance predictor. We show the actual and predicted accuracy of random sampling and our active sampling in Figure \ref{scattering}, where the MSE between the actual and predicted accuracy is also demonstrated. The predicted accuracy is generally closed to the actual one across the datasets, which shows the effectiveness of the performance predictor for automated model compression. Meanwhile, our active sampling strategy chooses the uncertain compression policies that provide informative supervision for performance predictor learning, so that the predicted accuracy is more precise compared with the random sampling strategy. Since the performance variance for different quantization and pruning strategies on largescale datasets is larger, and our active sampling strategy offers more benefits for the performance predictor learning on ImageNet. Although deeper architectures with large search space such as MobileNet-V2 and ResNet50 obtain higher MSE for their performance predictors, the active sampling policy is still capable of providing informative supervision for accurate performance predictor learning as the MSE is less than $5\times10^{-4}$.

\begin{table*}
	\scriptsize
	\centering
	\caption{The accuracy(\%) and BOPs(K) variance for Table 2 in the manuscript acquired by running experiments for $5$ times.}
	
	\label{ablation_objective_variance}
	\renewcommand\arraystretch{1.05}
	\begin{tabular}{c|c|cc|cc|cc|cc}
		\hline
		\multirow{2}{*}{Bar.}&\multirow{2}{*}{Gap.}&\multicolumn{2}{c|}{F$\backslash$M}&\multicolumn{2}{c|}{F\&M}&\multicolumn{2}{c|}{A$\backslash$M}&\multicolumn{2}{c}{A\&M}\\
		\cline{3-10}
		&& BOPs & Top-1& BOPs & Top-1& BOPs& Top-1& BOPs & Top-1\\
		\hline
		\multirow{2}{*}{$\times$}& $\times$& $193\pm1$ & $88.10\pm0.33$& $186\pm2$ & $88.42\pm0.25$& $198\pm1$ & $90.39\pm0.19$& $196\pm0$ & $90.57\pm0.23$\\
		\cline{2-10}
		&$\checkmark$& $185\pm3$ & $89.08\pm0.05$& $192\pm2$  & $89.17\pm0.10$& $196\pm1$  & $91.01\pm0.20$& $198\pm1$  & $91.12\pm0.35$\\
		\hline
		\multirow{2}{*}{$\checkmark$}& $\times$ & $194\pm0$ & $88.29\pm0.25$& $193\pm2$ & $88.66\pm0.31$& $190\pm3$ &  $90.72\pm0.28$& $193\pm1$ & $90.99\pm0.40$ \\
		\cline{2-10}
		& $\checkmark$ & $198\pm1$  & $89.22\pm0.25$& $196\pm1$ & $89.62\pm0.03$& $199\pm1$ & $91.07\pm0.13$& $200\pm0$ & $\bm{92.38}\pm0.17$ \\
		\hline	
	\end{tabular}
	\vspace{-0.5cm}
\end{table*}

\section*{C. Visualization of the Optimal Compression Policy}
We show the bitwidth of weights and activations and pruning ratio across different layers for image classification in Figure \ref{visualization}, where the computational cost constraint is $2.4$G and $0.2$G BOPs for compressing VGG-small and ResNet20 on CIFAR-10 and is $8$G, $33$G and $62$G BOPs for MobileNet-V2, ResNet18 and ResNet50 compression on ImageNet respectively.

Because sparse networks require precise weights and activations to maintain the representational capacity and increasing the bitwidth of layers with high pruning ratio only brings slight computational cost, the layers with high pruning ratio are usually assigned with large bitwidth in the optimal compression policy. The optimal compression policy searched via the state-of-the-art method DJPQ \citep{wang2020differentiable} only prunes the bottom layers since they sequentially prune the networks from bottom layers to top layers, while our SeerNet simultaneously optimizes the pruning strategy for all layers and preserves the informative channels with redundant channel removal.

For VGG-small and ResNet20 architectures trained on CIFAR-10, the bitwidth varies slightly across different layers. Meanwhile, the pruning ratio is high and the bitwidth is low for all layers, which means significant over-parameterization for both network architectures on CIFAR-10. The large bitwidth and low pruning ratio in the optimal compression policy for MobileNet-V2 indicates that the compact architecture is hard to be further compressed without sizable accuracy drop. On the contrary, ResNet50 is compressed with extremely low bitwidth, which demonstrates the significant redundancy. Moreover, the bitwidth of activations is usually larger than that of weights, which depicts that the model accuracy is more sensitive to activation quantization than weight quantization.

\section*{D. Implementation Details in Section 4.2.2}
To validate the effectiveness of the presented performance predictor, we utilized the reinforcement learning and evolutionary algorithms to search the optimal compression policy via our performance predictor. For reinforcement learning, we leveraged the deep deterministic policy gradient (DDPG) \citep{lillicrap2015continuous}. Compared with \citep{wang2019haq}, we added an extra state $p_k$ representing the pruning ratio for the $k_{th}$ convolutional layer in the state space and supplemented an extra action $a_k^p$ to sample the pruning ratio of compression strategy in the action space. The accuracy of the compressed model applied in the reward function was obtained via our performance predictor. Other implementation details were the same as those in \cite{wang2019haq}. For evolutionary algorithms, we followed the same implementation details in \cite{wang2020apq} to search the optimal compression policy except that we deleted the network architecture components for each candidate. The accuracy of each candidate applied in the fitness function was acquired via our performance predictor. We imposed the resource constraint by limiting the BOPs of the compressed models during the search process.

\begin{table}
	\renewcommand\arraystretch{1.05}
	\scriptsize
	\centering
	\caption{The accuracy(\%), MACs(K), BOPs(G) variance for Table 3 in the manuscript acquired by running experiments for $5$ times. }
	\label{CIFAR-10_variance}
	\begin{tabular}{c|ccc}	
		\hline
		Backbone &  MACs & BOPs  & Acc.\\
		\hline\hline
		\multirow{3}{*}{VGG-small}& $465\pm10$ & $7.37\pm0.03$ & $92.95\pm0.10$ \\
		& $332\pm12$ & $3.85\pm0.10$  & $92.69\pm0.26$ \\
		& $231\pm9$ & $2.29\pm0.07$ & $92.54\pm0.18$\\
		\hline
		\multirow{3}{*}{ResNet20} &$38\pm1$ & $0.63\pm0.02$ & $92.58\pm0.35$\\	
		&$31\pm2$ & $0.36\pm0.02$ & $92.35\pm0.24$\\
		&$37\pm2$ & $0.20\pm0$ & $92.38\pm0.17$\\
		\hline		
	\end{tabular}
	\vspace{-0.3cm}
\end{table}

\begin{table}
	\renewcommand\arraystretch{1.05}
	\scriptsize
	\centering
	\caption{The top-1 accuracy(\%), MACs(G), BOPs(G) variance for Table 4 in the manuscript acquired by running experiments for $5$ times. }
	\label{ImageNet_variance}
	\begin{tabular}{c|ccc}	
		\hline
		Backbone &  MACs & BOPs  & Top-1\\
		\hline\hline
		\multirow{3}{*}{MobileNet-V2}& $0.27\pm0.05$ & $10.93\pm0.05$ & $71.22\pm0.27$ \\
		& $0.22\pm0$ & $7.67\pm0.05$  & $70.78\pm0.07$ \\
		& $0.20\pm0.01$ & $4.90\pm0.02$ & $70.38\pm0.19$\\
		\hline
		\multirow{3}{*}{ResNet18} &$1.35\pm0.08$ & $55.98\pm0.99$ & $69.65\pm0.19$\\	
		&$1.25\pm0.06$ & $31.28\pm0.79$ & $69.15\pm0.14$\\
		&$0.68\pm0.07$ & $7.40\pm0.39$ & $67.52\pm0.27$\\
		\hline
		\multirow{3}{*}{ResNet50} &$3.25\pm0.17$ & $92.38\pm0.21$ & $76.50\pm0.30$\\	
		&$3.03\pm0.07$ & $60.06\pm0.20$ & $76.28\pm0.28$\\
		&$2.10\pm0.10$ & $31.19\pm0.52$ & $75.95\pm0.12$\\
		\hline		
	\end{tabular}
	\vspace{-0.3cm}
\end{table}

\begin{table}
	\renewcommand\arraystretch{1.05}
	\scriptsize
	\centering
	\caption{The mAP(\%), MACs(G), BOPs(G) variance for Table 5 in the manuscript acquired by running experiments for $5$ times. }
	\label{VOC_variance}
	\begin{tabular}{c|ccc}	
		\hline
		Backbone &  MACs & BOPs  & mAP\\
		\hline\hline
		\multirow{3}{*}{VGG16}& $18.08\pm0.23$ & $772.12\pm5.33$ & $69.5\pm0.2$ \\
		& $14.93\pm0.45$ & $629.78\pm10.33$  & $68.7\pm0.3$ \\
		& $14.01\pm0.69$ & $440.15\pm3.78$ & $66.9\pm0.3$\\
		\hline
		\multirow{3}{*}{ResNet18} &$19.00\pm0.15$ & $542.17\pm7.90$ & $72.9\pm0.2$\\	
		&$14.89\pm0.22$ & $338.13\pm6.89$ & $72.1\pm0.2$\\
		&$13.03\pm0.46$ & $295.29\pm12.23$ & $71.2\pm0.7$\\
		\hline		
	\end{tabular}
	\vspace{-0cm}
\end{table}

\begin{table}
	\renewcommand\arraystretch{1.05}
	\scriptsize
	\centering
	\caption{The mAP(\%), MACs(G), BOPs(G) variance for Table 6 in the manuscript acquired by running experiments for $5$ times. }
	\label{COCO_variance}
	\begin{tabular}{c|ccc}	
		\hline
		Backbone &  MACs & BOPs  & mAP\\
		\hline\hline
		\multirow{3}{*}{VGG16}& $20.92\pm0.86$ & $797.16\pm8.27$ & $22.6\pm0.1$ \\
		& $19.52\pm0.59$ & $558.18\pm7.10$  & $22.2\pm0.3$ \\
		& $15.59\pm0.19$ & $425.10\pm2.23$ & $21.3\pm0$\\
		\hline
		\multirow{3}{*}{ResNet18} &$19.99\pm0.15$ & $482.35\pm16.72$ & $26.8\pm0.1$\\	
		&$14.29\pm0.11$ & $318.90\pm8.72$ & $25.3\pm0.3$\\
		&$12.69\pm0.39$ & $289.75\pm4.50$ & $25.1\pm0$\\
		\hline		
	\end{tabular}
	\vspace{-0.3cm}
\end{table}

\section*{E. Performance Variance of SeerNet}
In order to show the performance variance of our SeerNet, we run SeerNet for $5$ times including compression policy search and backbone training for results in Table 1-6. We report the mean and standard deviation of accuracies and computational complexity due to the variation, while the search cost and training cost are almost the same for each time. We leveraged the same complexity budget as that in Table 1-6 of the manuscript, and Table 7-12 show the experimental results with mean and standard deviation.

\begin{table}[t]
	\renewcommand\arraystretch{1.05}
	\scriptsize
	\centering
	\caption{The MACs(G), BOPs(G), top-1 classification accuracy and search cost on ImageNet with pruning-only methods in MobileNet-V2. The search cost is demonstrated by GPU hours, where $N$ means the number of deployment scenarios. The number in the bracket of search cost for mixed-precision quantization demonstrates the break-even point of baseline methods whose search cost is higher than our SeerNet.}
	\vspace{0.1cm}              
	\label{ImageNet_pruning}
	\begin{tabular}{p{1.5cm}<{\centering}|p{0.55cm}<{\centering}p{0.55cm}<{\centering}p{0.55cm}<{\centering}p{0.75cm}<{\centering}|p{1.5cm}<{\centering}}	
		\hline
		Methods & MACs & BOPs & Comp. & Top-1 & Cost\\
		\hline
		Baseline & $0.33$ & $337.9$ &$-$& $71.72$ &$-$\\		
		AMC &$0.23$ &$236.5$ & $1.43$&$70.90$&$~~62.3N (13)$\\
		NetAdapt &$0.22$ &$225.3$ & $1.50$&$70.80$&$~~95.6N (8)$\\
		MetaPruning &$0.22$ &$222.2$ & $1.52$&$71.20$&$900$+$0.31N(1)$\\
		SeerNet & $0.22$ &$227.6$ & $1.48$& $71.52$ & $~750$+$0.002N$ \\
		\hline
	\end{tabular}
	%\vspace{0.35cm}
	\vspace{-0.3cm}
\end{table} 

\begin{table}[t]
	\renewcommand\arraystretch{1.05}
	\scriptsize
	\centering
	\caption{The MACs(G), BOPs(G), top-1 classification accuracy and search cost on ImageNet with quantization-only methods in MobileNet-V2, ResNet18 and ResNet50.}
	\vspace{0.1cm}              
	\label{ImageNet_quantization}
	\begin{tabular}{p{1.8cm}<{\centering}|p{1cm}<{\centering}|p{0.55cm}<{\centering}p{0.55cm}<{\centering}p{0.75cm}<{\centering}|p{1.3cm}<{\centering}}	
		\hline
		Backbone & Methods & MACs & BOPs & Top-1 & Cost\\
		\hline
		\multirow{3}{*}{MobileNet-V2}&Baseline & $0.33$ & $337.9$ &  $71.72$ &$-$\\		
		&HAQ &$0.33$ &$8.25$ & $69.45$&$~~51.1N (15)$\\
		&SeerNet & $0.33$ & $8.08$  & $70.64$ & $750$+$0.003N$ \\
		\hline
		\multirow{3}{*}{ResNet18}&Baseline &$1.81$ & $1853.4$& $69.74$ &$-$\\
		&	HAWQ & $1.81$&$34.00$&$68.45$&$~~22.7N(23)$\\	
		&	SeerNet & $1.81$ &$33.16$ & $69.38$ & $500$+$0.002N$\\
		\hline
		\multirow{3}{*}{ResNet50}&Baseline &$3.86$ & $3952.6$ & $76.40$ &$-$\\
		&	HAWQ  &$3.86$&$61.29$ &$75.48$&$~~34.5N (28)$\\
		&	SeerNet & $3.86$ &$58.49$ & $76.25$ & $950$+$0.003N$\\
		\hline
	\end{tabular}
	%\vspace{0.35cm}
	\vspace{-0cm}
\end{table}

\section*{F. Performance of Pruning-only and Quantization-only Strategies}
In this section, we evaluate SeerNet in the experimental settings with pruning-only and quantization-only strategies, where the backbone networks are only compressed by pruning and quantization policies in the above settings respectively. We implemented SeerNet following the details introduced in Section 4.1 of the manuscript except for the modifications that the compression policy sampling for performance predictor training only contains pruning or quantization for the two settings. The compared methods include AMC \citep{he2018amc}, NetAdapt \citep{yang2018netadapt}, MetaPruning \citep{liu2019metapruning} for pruning-only methods and contain HAQ \citep{wang2019haq} and HAWQ \citep{dong2019hawq} for quantization-only methods. For pruning-only strategies, since only one shared experimental setting exists in AMC, NetAdapt and MetaPruning which employs MobileNet-V2 with $0.22$G MACs for evaluation, we also assign the similar complexity constraint for fair comparison. For quantization-only strategies, we compare SeerNet with HAQ and HAWQ with MobileNet-V2, ResNet18 and ResNet50.  Table \ref{ImageNet_pruning} and \ref{ImageNet_quantization} illustrate the results for pruning-only and quantization-only strategies respectively, where our SeerNet still outperforms the baseline methods by a sizable margin with much less marginal search cost.

\begin{figure}[t]
	\centering
	\includegraphics[height=6.6cm, width=9cm]{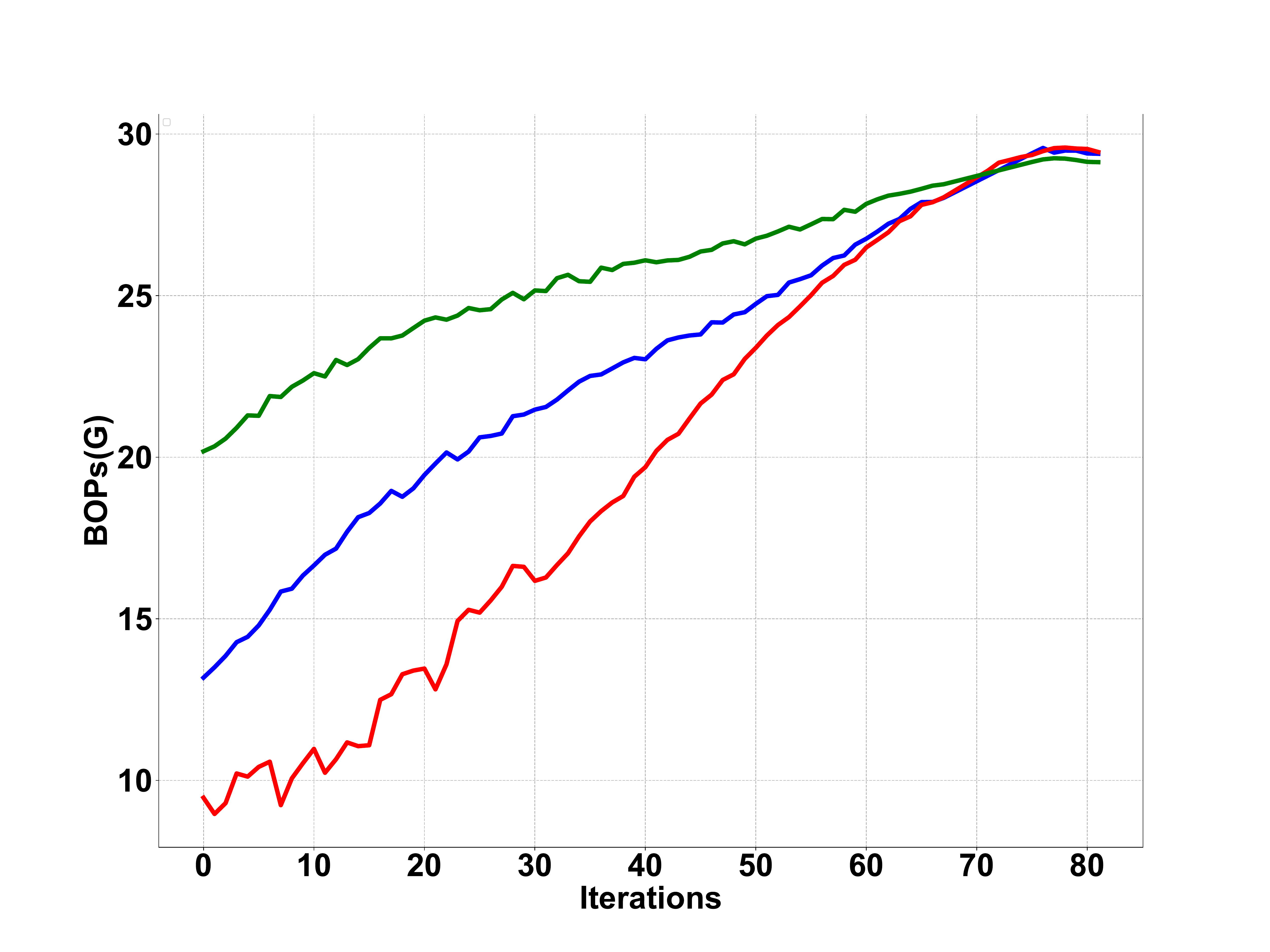}
	\caption{Several examples w.r.t. the compression policy complexity during the optimization with different policy initializations, where no optimization paths stop early because of exceeding the computational cost budget. Different colors represent various initializations, where the BOPs constraint is $30$G.}   
	\label{complexity_optimization}
	\vspace{0.3cm}
	%\vspace{-0.1cm}
\end{figure}

\section*{G. Visualization of Policy Optimization}
The presented barrier complexity loss in the compression policy optimization is amplified significantly for model complexity approaching the cost budget, which strictly limits the acquired pruning and quantization strategies within the complexity constraint. The complexity of the pruning and quantization policy in the optimization path usually keeps a margin from the computational cost budget. In order to empirically demonstrate the effectiveness of the barrier complexity loss, we implemented compression policy search for $50$ times with different initialization, and the compression policies with complexity higher than the budget were never observed during the compression policy optimization. Figure \ref{complexity_optimization} shows several examples w.r.t. the compression policy complexity during the optimization with different policy initializations, where no optimization paths stop early because of exceeding the computational cost budget.

\section*{H. Comparison with The Random Selection Baseline Method}
To show the effectiveness of our search method, we conducted experiments to compare our SeerNet with random selection baseline method (RS) with ResNet18 on ImageNet. The pipeline of RS is demonstrated as follows: (a) randomly sampling $k$ compression policies that satisfy the BOPs constraint, (b) exhaustively evaluating the acquired lightweight architectures, (c) selecting the one with the highest accuracy. With the same complexity constraint in Table 4 of the manuscript, we randomly sample 5 compression strategies that satisfy the BOPs budget for each random selection. Table \ref{RS} demonstrates the results, where the BOPs of RS is far from the budget and underperforms our SeerNet by a large margin regarding the accuracy. 

\begin{table}[t]
	\renewcommand\arraystretch{1.16}
	\scriptsize
	\centering
	\caption{The BOPs(G), top-1 classification accuracy and search cost on ImageNet with random search baseline and our SeerNet in ResNet18.}
	\vspace{0.1cm}              
	\label{RS}
	\begin{tabular}{p{1.1cm}<{\centering}|p{0.6cm}<{\centering}p{0.55cm}<{\centering}p{0.55cm}<{\centering}p{0.55cm}<{\centering}p{0.75cm}<{\centering}|p{1.3cm}<{\centering}}	
		\hline
		Methods & W/A & MACs & BOPs & Comp. & Top-1 & Cost\\
		\hline\hline
		\multicolumn{7}{c}{ResNet18}\\
		\hline
		Baseline& $32/32$ &$1.81$ & $1853.4$& $-$ & $69.74$ &$-$\\
		\hline
		RS &mixed &$1.35$& $53.51$& $34.64$& $66.83$  & $~~62.1N(9)$\\
		SeerNet& mixed & $1.37$ &$56.94$ & $32.55$ & $69.72$ & $500$+$0.003N$\\
		\hline
		RS & mixed & $1.18$&$28.75$&$54.51$&$67.00$&$~~62.1N(9)$\\	
		SeerNet& mixed & $1.22$ &$31.84$ & $58.21$ & $69.48$ & $500$+$0.004N$\\
		\hline
		RS& mixed &$0.69$ &$6.89$& $267.00$&$65.72$&$~~62.1N(9)$\\
		SeerNet& mixed & $0.70$ &$7.19$ & $257.71$ & $67.84$ & $500$+$0.004N$\\
		\hline
	\end{tabular}
	%\vspace{0.35cm}
	\vspace{0.3cm}
\end{table}

%\begin{acknowledgements}
%\end{acknowledgements}


{\footnotesize
	\bibliographystyle{plainnat}
	\bibliography{egbib}

\begin{thebibliography}{84}
\providecommand{\natexlab}[1]{#1}
\providecommand{\url}[1]{\texttt{#1}}
\expandafter\ifx\csname urlstyle\endcsname\relax
  \providecommand{\doi}[1]{doi: #1}\else
  \providecommand{\doi}{doi: \begingroup \urlstyle{rm}\Url}\fi

\bibitem[Abbasnejad et~al.(2020)Abbasnejad, Teney, Parvaneh, Shi, and
  Hengel]{abbasnejad2020counterfactual}
Ehsan Abbasnejad, Damien Teney, Amin Parvaneh, Javen Shi, and Anton van~den
  Hengel.
\newblock Counterfactual vision and language learning.
\newblock In \emph{CVPR}, pages 10044--10054, 2020.

\bibitem[Balcan et~al.(2007)Balcan, Broder, and Zhang]{balcan2007margin}
Maria-Florina Balcan, Andrei Broder, and Tong Zhang.
\newblock Margin based active learning.
\newblock In \emph{COLT}, pages 35--50, 2007.

\bibitem[Bell et~al.(2016)Bell, Lawrence~Zitnick, Bala, and
  Girshick]{bell2016inside}
Sean Bell, C~Lawrence~Zitnick, Kavita Bala, and Ross Girshick.
\newblock Inside-outside net: Detecting objects in context with skip pooling
  and recurrent neural networks.
\newblock In \emph{CVPR}, pages 2874--2883, 2016.

\bibitem[Beluch et~al.(2018)Beluch, Genewein, N{\"u}rnberger, and
  K{\"o}hler]{beluch2018power}
William~H Beluch, Tim Genewein, Andreas N{\"u}rnberger, and Jan~M K{\"o}hler.
\newblock The power of ensembles for active learning in image classification.
\newblock In \emph{CVPR}, pages 9368--9377, 2018.

\bibitem[Bethge et~al.(2020)Bethge, Bartz, Yang, Chen, and
  Meinel]{bethge2020meliusnet}
Joseph Bethge, Christian Bartz, Haojin Yang, Ying Chen, and Christoph Meinel.
\newblock Meliusnet: Can binary neural networks achieve mobilenet-level
  accuracy?
\newblock \emph{arXiv preprint arXiv:2001.05936}, 2020.

\bibitem[Bulat and Tzimiropoulos(2021)]{bulat2021bit}
Adrian Bulat and Georgios Tzimiropoulos.
\newblock Bit-mixer: Mixed-precision networks with runtime bit-width selection.
\newblock In \emph{ICCV}, pages 5188--5197, 2021.

\bibitem[Cai et~al.(2019)Cai, Gan, Wang, Zhang, and Han]{cai2019once}
Han Cai, Chuang Gan, Tianzhe Wang, Zhekai Zhang, and Song Han.
\newblock Once-for-all: Train one network and specialize it for efficient
  deployment.
\newblock \emph{arXiv preprint arXiv:1908.09791}, 2019.

\bibitem[Cai and Vasconcelos(2020)]{cai2020rethinking}
Zhaowei Cai and Nuno Vasconcelos.
\newblock Rethinking differentiable search for mixed-precision neural networks.
\newblock In \emph{CVPR}, pages 2349--2358, 2020.

\bibitem[Chen et~al.(2017)Chen, Choi, Yu, Han, and
  Chandraker]{chen2017learning}
Guobin Chen, Wongun Choi, Xiang Yu, Tony Han, and Manmohan Chandraker.
\newblock Learning efficient object detection models with knowledge
  distillation.
\newblock In \emph{NIPS}, pages 742--751, 2017.

\bibitem[Choi et~al.(2018)Choi, Wang, Venkataramani, Chuang, Srinivasan, and
  Gopalakrishnan]{choi2018pact}
Jungwook Choi, Zhuo Wang, Swagath Venkataramani, Pierce I-Jen Chuang,
  Vijayalakshmi Srinivasan, and Kailash Gopalakrishnan.
\newblock Pact: Parameterized clipping activation for quantized neural
  networks.
\newblock \emph{arXiv preprint arXiv:1805.06085}, 2018.

\bibitem[Dai et~al.(2019)Dai, Zhang, Wu, Yin, Sun, Wang, Dukhan, Hu, Wu, Jia,
  et~al.]{dai2019chamnet}
Xiaoliang Dai, Peizhao Zhang, Bichen Wu, Hongxu Yin, Fei Sun, Yanghan Wang,
  Marat Dukhan, Yunqing Hu, Yiming Wu, Yangqing Jia, et~al.
\newblock Chamnet: Towards efficient network design through platform-aware
  model adaptation.
\newblock In \emph{CVPR}, pages 11398--11407, 2019.

\bibitem[Deng et~al.(2009)Deng, Dong, Socher, Li, Li, and
  Fei-Fei]{deng2009imagenet}
Jia Deng, Wei Dong, Richard Socher, Li-Jia Li, Kai Li, and Li~Fei-Fei.
\newblock Imagenet: A large-scale hierarchical image database.
\newblock In \emph{CVPR}, pages 248--255, 2009.

\bibitem[Denil et~al.(2013)Denil, Shakibi, Dinh, De~Freitas,
  et~al.]{denil2013predicting}
Misha Denil, Babak Shakibi, Laurent Dinh, Nando De~Freitas, et~al.
\newblock Predicting parameters in deep learning.
\newblock In \emph{NIPS}, pages 2148--2156, 2013.

\bibitem[Dong et~al.(2018)Dong, Liao, Pang, Su, Zhu, Hu, and
  Li]{dong2018boosting}
Yinpeng Dong, Fangzhou Liao, Tianyu Pang, Hang Su, Jun Zhu, Xiaolin Hu, and
  Jianguo Li.
\newblock Boosting adversarial attacks with momentum.
\newblock In \emph{CVPR}, pages 9185--9193, 2018.

\bibitem[Dong et~al.(2019)Dong, Yao, Gholami, Mahoney, and
  Keutzer]{dong2019hawq}
Zhen Dong, Zhewei Yao, Amir Gholami, Michael~W Mahoney, and Kurt Keutzer.
\newblock Hawq: Hessian aware quantization of neural networks with
  mixed-precision.
\newblock In \emph{ICCV}, pages 293--302, 2019.

\bibitem[Duch and Korczak(1998)]{duch1998optimization}
W{\l}odzis{\l}aw Duch and Jerzy Korczak.
\newblock Optimization and global minimization methods suitable for neural
  networks.
\newblock \emph{Neural computing surveys}, 2:\penalty0 163--212, 1998.

\bibitem[Erin~Liong et~al.(2015)Erin~Liong, Lu, Wang, Moulin, and
  Zhou]{erin2015deep}
Venice Erin~Liong, Jiwen Lu, Gang Wang, Pierre Moulin, and Jie Zhou.
\newblock Deep hashing for compact binary codes learning.
\newblock In \emph{CVPR}, pages 2475--2483, 2015.

\bibitem[Esser et~al.(2019)Esser, McKinstry, Bablani, Appuswamy, and
  Modha]{esser2019learned}
Steven~K Esser, Jeffrey~L McKinstry, Deepika Bablani, Rathinakumar Appuswamy,
  and Dharmendra~S Modha.
\newblock Learned step size quantization.
\newblock \emph{arXiv preprint arXiv:1902.08153}, 2019.

\bibitem[Everingham et~al.(2010)Everingham, Van~Gool, Williams, Winn, and
  Zisserman]{everingham2010pascal}
Mark Everingham, Luc Van~Gool, Christopher~KI Williams, John Winn, and Andrew
  Zisserman.
\newblock The pascal visual object classes (voc) challenge.
\newblock \emph{IJCV}, 88\penalty0 (2):\penalty0 303--338, 2010.

\bibitem[Feichtenhofer et~al.(2019)Feichtenhofer, Fan, Malik, and
  He]{feichtenhofer2019slowfast}
Christoph Feichtenhofer, Haoqi Fan, Jitendra Malik, and Kaiming He.
\newblock Slowfast networks for video recognition.
\newblock In \emph{ICCV}, pages 6202--6211, 2019.

\bibitem[Finlay et~al.(2019)Finlay, Pooladian, and
  Oberman]{finlay2019logbarrier}
Chris Finlay, Aram-Alexandre Pooladian, and Adam Oberman.
\newblock The logbarrier adversarial attack: making effective use of decision
  boundary information.
\newblock In \emph{ICCV}, pages 4862--4870, 2019.

\bibitem[Gal et~al.(2017)Gal, Islam, and Ghahramani]{gal2017deep}
Yarin Gal, Riashat Islam, and Zoubin Ghahramani.
\newblock Deep bayesian active learning with image data.
\newblock \emph{arXiv preprint arXiv:1703.02910}, 2017.

\bibitem[Gong et~al.(2019)Gong, Liu, Jiang, Li, Hu, Lin, Yu, and
  Yan]{gong2019differentiable}
Ruihao Gong, Xianglong Liu, Shenghu Jiang, Tianxiang Li, Peng Hu, Jiazhen Lin,
  Fengwei Yu, and Junjie Yan.
\newblock Differentiable soft quantization: Bridging full-precision and low-bit
  neural networks.
\newblock \emph{arXiv preprint arXiv:1908.05033}, 2019.

\bibitem[Goyal et~al.(2019)Goyal, Wu, Ernst, Batra, Parikh, and
  Lee]{goyal2019counterfactual}
Yash Goyal, Ziyan Wu, Jan Ernst, Dhruv Batra, Devi Parikh, and Stefan Lee.
\newblock Counterfactual visual explanations.
\newblock \emph{arXiv preprint arXiv:1904.07451}, 2019.

\bibitem[Habi et~al.(2020)Habi, Jennings, and Netzer]{habi2020hmq}
Hai~Victor Habi, Roy~H Jennings, and Arnon Netzer.
\newblock Hmq: Hardware friendly mixed precision quantization block for cnns.
\newblock \emph{arXiv preprint arXiv:2007.09952}, 2020.

\bibitem[Han et~al.(2015{\natexlab{a}})Han, Mao, and Dally]{han2015deep}
Song Han, Huizi Mao, and William~J Dally.
\newblock Deep compression: Compressing deep neural networks with pruning,
  trained quantization and huffman coding.
\newblock \emph{arXiv preprint arXiv:1510.00149}, 2015{\natexlab{a}}.

\bibitem[Han et~al.(2015{\natexlab{b}})Han, Pool, Tran, and
  Dally]{han2015learning}
Song Han, Jeff Pool, John Tran, and William Dally.
\newblock Learning both weights and connections for efficient neural network.
\newblock In \emph{NIPS}, pages 1135--1143, 2015{\natexlab{b}}.

\bibitem[He et~al.(2016)He, Zhang, Ren, and Sun]{he2016deep}
Kaiming He, Xiangyu Zhang, Shaoqing Ren, and Jian Sun.
\newblock Deep residual learning for image recognition.
\newblock In \emph{CVPR}, pages 770--778, 2016.

\bibitem[He et~al.(2018{\natexlab{a}})He, Kang, Dong, Fu, and Yang]{he2018soft}
Yang He, Guoliang Kang, Xuanyi Dong, Yanwei Fu, and Yi~Yang.
\newblock Soft filter pruning for accelerating deep convolutional neural
  networks.
\newblock \emph{arXiv preprint arXiv:1808.06866}, 2018{\natexlab{a}}.

\bibitem[He et~al.(2017)He, Zhang, and Sun]{he2017channel}
Yihui He, Xiangyu Zhang, and Jian Sun.
\newblock Channel pruning for accelerating very deep neural networks.
\newblock In \emph{ICCV}, pages 1389--1397, 2017.

\bibitem[He et~al.(2018{\natexlab{b}})He, Lin, Liu, Wang, Li, and
  Han]{he2018amc}
Yihui He, Ji~Lin, Zhijian Liu, Hanrui Wang, Li-Jia Li, and Song Han.
\newblock Amc: Automl for model compression and acceleration on mobile devices.
\newblock In \emph{ECCV}, pages 784--800, 2018{\natexlab{b}}.

\bibitem[Howard et~al.(2017)Howard, Zhu, Chen, Kalenichenko, Wang, Weyand,
  Andreetto, and Adam]{howard2017mobilenets}
Andrew~G Howard, Menglong Zhu, Bo~Chen, Dmitry Kalenichenko, Weijun Wang,
  Tobias Weyand, Marco Andreetto, and Hartwig Adam.
\newblock Mobilenets: Efficient convolutional neural networks for mobile vision
  applications.
\newblock \emph{arXiv preprint arXiv:1704.04861}, 2017.

\bibitem[Hubara et~al.(2016)Hubara, Courbariaux, Soudry, El-Yaniv, and
  Bengio]{hubara2016binarized}
Itay Hubara, Matthieu Courbariaux, Daniel Soudry, Ran El-Yaniv, and Yoshua
  Bengio.
\newblock Binarized neural networks.
\newblock In \emph{NIPS}, pages 4107--4115, 2016.

\bibitem[Jin et~al.(2020)Jin, Yang, and Liao]{jin2020adabits}
Qing Jin, Linjie Yang, and Zhenyu Liao.
\newblock Adabits: Neural network quantization with adaptive bit-widths.
\newblock In \emph{CVPR}, pages 2146--2156, 2020.

\bibitem[Joshi et~al.(2009)Joshi, Porikli, and
  Papanikolopoulos]{joshi2009multi}
Ajay~J Joshi, Fatih Porikli, and Nikolaos Papanikolopoulos.
\newblock Multi-class active learning for image classification.
\newblock In \emph{CVPR}, pages 2372--2379, 2009.

\bibitem[Kingma and Ba(2014)]{kingma2014adam}
Diederik~P Kingma and Jimmy Ba.
\newblock Adam: A method for stochastic optimization.
\newblock \emph{arXiv preprint arXiv:1412.6980}, 2014.

\bibitem[Krizhevsky and Hinton(2009)]{krizhevsky2009learning}
Alex Krizhevsky and Geoffrey Hinton.
\newblock Learning multiple layers of features from tiny images.
\newblock 2009.

\bibitem[Li et~al.(2016)Li, Kadav, Durdanovic, Samet, and Graf]{li2016pruning}
Hao Li, Asim Kadav, Igor Durdanovic, Hanan Samet, and Hans~Peter Graf.
\newblock Pruning filters for efficient convnets.
\newblock \emph{arXiv preprint arXiv:1608.08710}, 2016.

\bibitem[Li et~al.(2019{\natexlab{a}})Li, Qi, Wang, Ge, Li, Yue, and
  Sun]{li2019oicsr}
Jiashi Li, Qi~Qi, Jingyu Wang, Ce~Ge, Yujian Li, Zhangzhang Yue, and Haifeng
  Sun.
\newblock Oicsr: Out-in-channel sparsity regularization for compact deep neural
  networks.
\newblock In \emph{CVPR}, pages 7046--7055, 2019{\natexlab{a}}.

\bibitem[Li et~al.(2019{\natexlab{b}})Li, Wang, Liang, Qin, Yan, and
  Fan]{li2019fully}
Rundong Li, Yan Wang, Feng Liang, Hongwei Qin, Junjie Yan, and Rui Fan.
\newblock Fully quantized network for object detection.
\newblock In \emph{CVPR}, pages 2810--2819, 2019{\natexlab{b}}.

\bibitem[Li and Guo(2014)]{li2014multi}
Xin Li and Yuhong Guo.
\newblock Multi-level adaptive active learning for scene classification.
\newblock In \emph{ECCV}, pages 234--249, 2014.

\bibitem[Li et~al.(2020{\natexlab{a}})Li, Gu, Mayer, Gool, and
  Timofte]{li2020group}
Yawei Li, Shuhang Gu, Christoph Mayer, Luc~Van Gool, and Radu Timofte.
\newblock Group sparsity: The hinge between filter pruning and decomposition
  for network compression.
\newblock In \emph{CVPR}, pages 8018--8027, 2020{\natexlab{a}}.

\bibitem[Li et~al.(2020{\natexlab{b}})Li, Dong, and Wang]{li2019additive}
Yuhang Li, Xin Dong, and Wei Wang.
\newblock Additive powers-of-two quantization: A non-uniform discretization for
  neural networks.
\newblock \emph{ICLR}, 2020{\natexlab{b}}.

\bibitem[Lillicrap et~al.(2015)Lillicrap, Hunt, Pritzel, Heess, Erez, Tassa,
  Silver, and Wierstra]{lillicrap2015continuous}
Timothy~P Lillicrap, Jonathan~J Hunt, Alexander Pritzel, Nicolas Heess, Tom
  Erez, Yuval Tassa, David Silver, and Daan Wierstra.
\newblock Continuous control with deep reinforcement learning.
\newblock \emph{arXiv preprint arXiv:1509.02971}, 2015.

\bibitem[Lin et~al.(2014)Lin, Maire, Belongie, Hays, Perona, Ramanan,
  Doll{\'a}r, and Zitnick]{lin2014microsoft}
Tsung-Yi Lin, Michael Maire, Serge Belongie, James Hays, Pietro Perona, Deva
  Ramanan, Piotr Doll{\'a}r, and C~Lawrence Zitnick.
\newblock Microsoft coco: Common objects in context.
\newblock In \emph{ECCV}, pages 740--755, 2014.

\bibitem[Liu et~al.(2015)Liu, Wang, Foroosh, Tappen, and Pensky]{liu2015sparse}
Baoyuan Liu, Min Wang, Hassan Foroosh, Marshall Tappen, and Marianna Pensky.
\newblock Sparse convolutional neural networks.
\newblock In \emph{CVPR}, pages 806--814, 2015.

\bibitem[Liu et~al.(2016)Liu, Anguelov, Erhan, Szegedy, Reed, Fu, and
  Berg]{liu2016ssd}
Wei Liu, Dragomir Anguelov, Dumitru Erhan, Christian Szegedy, Scott Reed,
  Cheng-Yang Fu, and Alexander~C Berg.
\newblock Ssd: Single shot multibox detector.
\newblock In \emph{ECCV}, pages 21--37, 2016.

\bibitem[Liu et~al.(2018{\natexlab{a}})Liu, Wu, Luo, Yang, Liu, and
  Cheng]{liu2018bi}
Zechun Liu, Baoyuan Wu, Wenhan Luo, Xin Yang, Wei Liu, and Kwang-Ting Cheng.
\newblock Bi-real net: Enhancing the performance of 1-bit cnns with improved
  representational capability and advanced training algorithm.
\newblock In \emph{ECCV}, pages 722--737, 2018{\natexlab{a}}.

\bibitem[Liu et~al.(2019)Liu, Mu, Zhang, Guo, Yang, Cheng, and
  Sun]{liu2019metapruning}
Zechun Liu, Haoyuan Mu, Xiangyu Zhang, Zichao Guo, Xin Yang, Kwang-Ting Cheng,
  and Jian Sun.
\newblock Metapruning: Meta learning for automatic neural network channel
  pruning.
\newblock In \emph{ICCV}, pages 3296--3305, 2019.

\bibitem[Liu et~al.(2018{\natexlab{b}})Liu, Sun, Zhou, Huang, and
  Darrell]{liu2018rethinking}
Zhuang Liu, Mingjie Sun, Tinghui Zhou, Gao Huang, and Trevor Darrell.
\newblock Rethinking the value of network pruning.
\newblock In \emph{ICLR}, 2018{\natexlab{b}}.

\bibitem[Lou et~al.(2019)Lou, Guo, Kim, Liu, and Jiang]{lou2019autoq}
Qian Lou, Feng Guo, Minje Kim, Lantao Liu, and Lei Jiang.
\newblock Autoq: Automated kernel-wise neural network quantization.
\newblock In \emph{ICLR}, 2019.

\bibitem[Louizos et~al.(2017)Louizos, Welling, and Kingma]{louizos2017learning}
Christos Louizos, Max Welling, and Diederik~P Kingma.
\newblock Learning sparse neural networks through $ l\_0 $ regularization.
\newblock \emph{arXiv preprint arXiv:1712.01312}, 2017.

\bibitem[Louizos et~al.(2018)Louizos, Reisser, Blankevoort, Gavves, and
  Welling]{louizos2018relaxed}
Christos Louizos, Matthias Reisser, Tijmen Blankevoort, Efstratios Gavves, and
  Max Welling.
\newblock Relaxed quantization for discretized neural networks.
\newblock \emph{arXiv preprint arXiv:1810.01875}, 2018.

\bibitem[Luo et~al.(2013)Luo, Schwing, and Urtasun]{luo2013latent}
Wenjie Luo, Alex Schwing, and Raquel Urtasun.
\newblock Latent structured active learning.
\newblock \emph{NIPS}, 26:\penalty0 728--736, 2013.

\bibitem[Melville and Mooney(2004)]{melville2004diverse}
Prem Melville and Raymond~J Mooney.
\newblock Diverse ensembles for active learning.
\newblock In \emph{ICML}, page~74, 2004.

\bibitem[Molchanov et~al.(2016)Molchanov, Tyree, Karras, Aila, and
  Kautz]{molchanov2016pruning}
Pavlo Molchanov, Stephen Tyree, Tero Karras, Timo Aila, and Jan Kautz.
\newblock Pruning convolutional neural networks for resource efficient
  inference.
\newblock \emph{arXiv preprint arXiv:1611.06440}, 2016.

\bibitem[Molchanov et~al.(2019)Molchanov, Mallya, Tyree, Frosio, and
  Kautz]{molchanov2019importance}
Pavlo Molchanov, Arun Mallya, Stephen Tyree, Iuri Frosio, and Jan Kautz.
\newblock Importance estimation for neural network pruning.
\newblock In \emph{CVPR}, pages 11264--11272, 2019.

\bibitem[Peng et~al.(2019)Peng, Wu, Chen, and Huang]{peng2019collaborative}
Hanyu Peng, Jiaxiang Wu, Shifeng Chen, and Junzhou Huang.
\newblock Collaborative channel pruning for deep networks.
\newblock In \emph{ICML}, pages 5113--5122, 2019.

\bibitem[Phan et~al.(2019)Phan, Huynh, He, Savvides, and Shen]{phan2019mobinet}
Hai Phan, Dang Huynh, Yihui He, Marios Savvides, and Zhiqiang Shen.
\newblock Mobinet: A mobile binary network for image classification.
\newblock \emph{arXiv preprint arXiv:1907.12629}, 2019.

\bibitem[Qu et~al.(2020)Qu, Zhou, Cheng, and Thiele]{qu2020adaptive}
Zhongnan Qu, Zimu Zhou, Yun Cheng, and Lothar Thiele.
\newblock Adaptive loss-aware quantization for multi-bit networks.
\newblock In \emph{CVPR}, pages 7988--7997, 2020.

\bibitem[Rastegari et~al.(2016)Rastegari, Ordonez, Redmon, and
  Farhadi]{rastegari2016xnor}
Mohammad Rastegari, Vicente Ordonez, Joseph Redmon, and Ali Farhadi.
\newblock Xnor-net: Imagenet classification using binary convolutional neural
  networks.
\newblock In \emph{ECCV}, pages 525--542, 2016.

\bibitem[Ren et~al.(2015)Ren, He, Girshick, and Sun]{ren2015faster}
Shaoqing Ren, Kaiming He, Ross Girshick, and Jian Sun.
\newblock Faster r-cnn: Towards real-time object detection with region proposal
  networks.
\newblock In \emph{NIPS}, pages 91--99, 2015.

\bibitem[Sandler et~al.(2018)Sandler, Howard, Zhu, Zhmoginov, and
  Chen]{sandler2018mobilenetv2}
Mark Sandler, Andrew Howard, Menglong Zhu, Andrey Zhmoginov, and Liang-Chieh
  Chen.
\newblock Mobilenetv2: Inverted residuals and linear bottlenecks.
\newblock In \emph{CVPR}, pages 4510--4520, 2018.

\bibitem[Settles and Craven(2008)]{settles2008analysis}
Burr Settles and Mark Craven.
\newblock An analysis of active learning strategies for sequence labeling
  tasks.
\newblock In \emph{EMNLP}, pages 1070--1079, 2008.

\bibitem[Siddiqui et~al.(2020)Siddiqui, Valentin, and
  Nie{\ss}ner]{siddiqui2020viewal}
Yawar Siddiqui, Julien Valentin, and Matthias Nie{\ss}ner.
\newblock Viewal: Active learning with viewpoint entropy for semantic
  segmentation.
\newblock In \emph{CVPR}, pages 9433--9443, 2020.

\bibitem[Simonyan and Zisserman(2014)]{simonyan2014very}
Karen Simonyan and Andrew Zisserman.
\newblock Very deep convolutional networks for large-scale image recognition.
\newblock \emph{arXiv preprint arXiv:1409.1556}, 2014.

\bibitem[Sutskever et~al.(2013)Sutskever, Martens, Dahl, and
  Hinton]{sutskever2013importance}
Ilya Sutskever, James Martens, George Dahl, and Geoffrey Hinton.
\newblock On the importance of initialization and momentum in deep learning.
\newblock In \emph{ICML}, pages 1139--1147, 2013.

\bibitem[Uhlich et~al.(2019)Uhlich, Mauch, Yoshiyama, Cardinaux, Garcia,
  Tiedemann, Kemp, and Nakamura]{uhlich2019differentiable}
Stefan Uhlich, Lukas Mauch, Kazuki Yoshiyama, Fabien Cardinaux, Javier~Alonso
  Garcia, Stephen Tiedemann, Thomas Kemp, and Akira Nakamura.
\newblock Differentiable quantization of deep neural networks.
\newblock \emph{arXiv preprint arXiv:1905.11452}, 2019.

\bibitem[Vasisht et~al.(2014)Vasisht, Damianou, Varma, and
  Kapoor]{vasisht2014active}
Deepak Vasisht, Andreas Damianou, Manik Varma, and Ashish Kapoor.
\newblock Active learning for sparse bayesian multilabel classification.
\newblock In \emph{KDD}, pages 472--481, 2014.

\bibitem[Vijayanarasimhan and Grauman(2014)]{vijayanarasimhan2014large}
Sudheendra Vijayanarasimhan and Kristen Grauman.
\newblock Large-scale live active learning: Training object detectors with
  crawled data and crowds.
\newblock \emph{IJCV}, 108\penalty0 (1-2):\penalty0 97--114, 2014.

\bibitem[Wang et~al.(2019{\natexlab{a}})Wang, Liu, Lin, Lin, and
  Han]{wang2019haq}
Kuan Wang, Zhijian Liu, Yujun Lin, Ji~Lin, and Song Han.
\newblock Haq: Hardware-aware automated quantization with mixed precision.
\newblock In \emph{CVPR}, pages 8612--8620, 2019{\natexlab{a}}.

\bibitem[Wang et~al.(2020{\natexlab{a}})Wang, Wang, Cai, Lin, Liu, Wang, Lin,
  and Han]{wang2020apq}
Tianzhe Wang, Kuan Wang, Han Cai, Ji~Lin, Zhijian Liu, Hanrui Wang, Yujun Lin,
  and Song Han.
\newblock Apq: Joint search for network architecture, pruning and quantization
  policy.
\newblock In \emph{CVPR}, pages 2078--2087, 2020{\natexlab{a}}.

\bibitem[Wang et~al.(2019{\natexlab{b}})Wang, Song, Zhao, Shen, Zhao, Hoi, and
  Ling]{wang2019learning1}
Wenguan Wang, Hongmei Song, Shuyang Zhao, Jianbing Shen, Sanyuan Zhao,
  Steven~CH Hoi, and Haibin Ling.
\newblock Learning unsupervised video object segmentation through visual
  attention.
\newblock In \emph{CVPR}, pages 3064--3074, 2019{\natexlab{b}}.

\bibitem[Wang et~al.(2020{\natexlab{b}})Wang, Lu, and
  Blankevoort]{wang2020differentiable}
Ying Wang, Yadong Lu, and Tijmen Blankevoort.
\newblock Differentiable joint pruning and quantization for hardware
  efficiency.
\newblock In \emph{ECCV}, pages 259--277, 2020{\natexlab{b}}.

\bibitem[Wang et~al.(2020{\natexlab{c}})Wang, Zheng, Lu, and
  Zhou]{wang2020deep}
Ziwei Wang, Quan Zheng, Jiwen Lu, and Jie Zhou.
\newblock Deep hashing with active pairwise supervision.
\newblock In \emph{ECCV}, pages 522--538, 2020{\natexlab{c}}.

\bibitem[Wang et~al.(2021{\natexlab{a}})Wang, Lu, and Zhou]{wang2021learning}
Ziwei Wang, Jiwen Lu, and Jie Zhou.
\newblock Learning channel-wise interactions for binary convolutional neural
  networks.
\newblock \emph{TPAMI}, 43\penalty0 (10):\penalty0 3432--3445,
  2021{\natexlab{a}}.

\bibitem[Wang et~al.(2021{\natexlab{b}})Wang, Xiao, Lu, and
  Zhou]{wang2021generalizable}
Ziwei Wang, Han Xiao, Jiwen Lu, and Jie Zhou.
\newblock Generalizable mixed-precision quantization via attribution rank
  preservation.
\newblock In \emph{ICCV}, pages 5291--5300, 2021{\natexlab{b}}.

\bibitem[Wang et~al.(2022{\natexlab{a}})Wang, Lu, Wu, and
  Zhou]{wang2022learning}
Ziwei Wang, Jiwen Lu, Ziyi Wu, and Jie Zhou.
\newblock Learning efficient binarized object detectors with information
  compression.
\newblock \emph{TPAMI}, 44\penalty0 (6):\penalty0 3082--3095,
  2022{\natexlab{a}}.

\bibitem[Wang et~al.(2022{\natexlab{b}})Wang, Wang, Xu, Zhou, and
  Lu]{wang2022quantformer}
Ziwei Wang, Changyuan Wang, Xiuwei Xu, Jie Zhou, and Jiwen Lu.
\newblock Quantformer: Learning extremely low-precision vision transformers.
\newblock \emph{TPAMI}, pages 1--14, 2022{\natexlab{b}}.
\newblock \doi{10.1109/TPAMI.2022.3229313}.

\bibitem[Wen et~al.(2020)Wen, Liu, Chen, Li, Bender, and
  Kindermans]{wen2020neural}
Wei Wen, Hanxiao Liu, Yiran Chen, Hai Li, Gabriel Bender, and Pieter-Jan
  Kindermans.
\newblock Neural predictor for neural architecture search.
\newblock In \emph{ECCV}, pages 660--676, 2020.

\bibitem[Wu et~al.(2022)Wu, Wang, Wei, Wei, and Yan]{wu2022smart}
Zhenyu Wu, Ziwei Wang, Zibu Wei, Yi~Wei, and Haibin Yan.
\newblock Smart explorer: Recognizing objects in dense clutter via interactive
  exploration.
\newblock In \emph{IROS}, pages 6600--6607, 2022.

\bibitem[Yang et~al.(2018)Yang, Howard, Chen, Zhang, Go, Sandler, Sze, and
  Adam]{yang2018netadapt}
Tien-Ju Yang, Andrew Howard, Bo~Chen, Xiao Zhang, Alec Go, Mark Sandler,
  Vivienne Sze, and Hartwig Adam.
\newblock Netadapt: Platform-aware neural network adaptation for mobile
  applications.
\newblock In \emph{ECCV}, pages 285--300, 2018.

\bibitem[Yu et~al.(2020)Yu, Han, Li, Shi, Cheng, and Fan]{yu2020search}
Haibao Yu, Qi~Han, Jianbo Li, Jianping Shi, Guangliang Cheng, and Bin Fan.
\newblock Search what you want: Barrier panelty nas for mixed precision
  quantization.
\newblock \emph{arXiv preprint arXiv:2007.10026}, 2020.

\bibitem[Zhang et~al.(2018)Zhang, Yang, Ye, and Hua]{zhang2018lq}
Dongqing Zhang, Jiaolong Yang, Dongqiangzi Ye, and Gang Hua.
\newblock Lq-nets: Learned quantization for highly accurate and compact deep
  neural networks.
\newblock In \emph{ECCV}, pages 365--382, 2018.

\end{thebibliography}
}

\end{document}